\documentclass{article}

\PassOptionsToPackage{numbers, compress}{natbib}
\usepackage[preprint]{neurips_2024}
\usepackage{graphicx}
\usepackage{caption}
\usepackage{subcaption}

\usepackage[utf8]{inputenc} % allow utf-8 input
\usepackage[T1]{fontenc}    % use 8-bit T1 fonts
\usepackage{hyperref}       % hyperlinks
\usepackage{url}            % simple URL typesetting
\usepackage{booktabs}       % professional-quality tables
\usepackage{amsfonts}       % blackboard math symbols
\usepackage{nicefrac}       % compact symbols for 1/2, etc.
\usepackage{microtype}      % microtypography
\usepackage{xcolor}         % colors

\usepackage{enumitem}
\usepackage{algorithm}
\usepackage[noend]{algpseudocode}
\usepackage{amsmath}
\usepackage{amssymb}
\usepackage{caption}
\usepackage{makecell}
\usepackage{multirow}
\usepackage{wrapfig}
\usepackage{graphicx}
\usepackage{subcaption}
\usepackage{pifont}

\usepackage{algorithm}
\usepackage[noend]{algpseudocode}
\usepackage{amsfonts}
\usepackage{amsmath}
\usepackage{amssymb}
\usepackage{booktabs}
\usepackage{caption}
\usepackage{cleveref}
\usepackage{enumitem}
\usepackage{float}
\usepackage{graphicx}
\usepackage{hyperref}
\usepackage{makecell}
\usepackage{multirow}
\usepackage{natbib}
\usepackage{subcaption}
\usepackage[table]{xcolor}
\usepackage{wrapfig}
\usepackage{xcolor}
\usepackage{pifont}
\usepackage{times}
\usepackage[many]{tcolorbox}  
\usepackage{latexsym}
\definecolor{cvprblue}{rgb}{0.21,0.49,0.74}

\usepackage{tikz}
\usepackage{xcolor}
\usepackage{newunicodechar}
\usepackage{pifont}
\newunicodechar{−}{\ensuremath{-}}

\newcommand{\rvqa}{{Q-Router}}

\newtcolorbox{boxK}[2][]{
    sharpish corners, % better drop shadow
    boxrule = 0pt,
    toprule = 4.5pt, % top rule weight
    enhanced,
    fuzzy shadow = {0pt}{-2pt}{-0.5pt}{0.5pt}{black!35}, % {xshift}{yshift}{offset}{step}{options}
    fontupper = \ttfamily\small, % Use a monospaced font (e.g., Courier)
    boxsep = 5pt, % padding inside the box
    left = 5pt, % left padding
    right = 5pt, % right padding
    top = 5pt, % top padding
    bottom = 5pt, % bottom padding
    #1                       % allow override of color per box
}

\title{Q-Router: Agentic Video Quality Assessment with Expert Model Routing and Artifact Localization}

\author{%
  Shuo Xing\thanks{Use footnote for providing further information
    about author (webpage, alternative address)---\emph{not} for acknowledging
    funding agencies.}, Chengyuan Qian \\
  Department of Computer Science\\
  Cranberry-Lemon University\\
  Pittsburgh, PA 15213 \\
  \texttt{hippo@cs.cranberry-lemon.edu} \\
}

\author{
\bf Shuo Xing$^1$,
Soumik Dey$^2$, 
Mingyang Wu$^1$, 
Ashirbad Mishra$^2$, 
Naveen Ravipati$^2$, \\ 
\bf Binbin Li$^2$,
Hansi Wu$^2$, 
Zhengzhong Tu$^1$\thanks{Corresponding author: \texttt{tzz@tamu.edu}}
\\
\\
$^1$Texas A\&M University  \quad
$^2$eBay Inc.
\\
}

\begin{document}

\maketitle

\begin{abstract}
Video quality assessment (VQA) is a fundamental computer vision task that aims to predict the perceptual quality of a given video in alignment with human judgments. Existing performant VQA models trained with direct score supervision suffer from \textbf{(1)} \emph{poor generalization} across diverse content and tasks, ranging from user-generated content (UGC), short-form videos, to AI-generated content (AIGC), \textbf{(2)} \emph{limited interpretability}, and \textbf{(3)} \emph{lack of extensibility} to novel use cases or content types. 
We propose \rvqa{}, an agentic framework for universal VQA with a multi-tier model routing system. \rvqa{} integrates a diverse set of expert models and employs vision–language models (VLMs) as real-time routers that dynamically reason then ensemble the most appropriate experts conditioned on the input video semantics. 
We build a multi-tiered routing system based on the computing budget, with the heaviest tier involving a specific spatiotemporal artifacts localization for interpretability.
This agentic design enables \rvqa{} to combine the complementary strengths of specialized experts, achieving both flexibility and robustness in delivering consistent performance across heterogeneous video sources and tasks.  Extensive experiments demonstrate that \rvqa{} matches or surpasses state-of-the-art VQA models on a variety of benchmarks, while substantially improving generalization and interpretability. Moreover, \rvqa{} excels on the quality-based question answering benchmark, Q-Bench-Video, highlighting its promise as a foundation for next-generation VQA systems. Finally, we show that \rvqa{} capably localize spatiotemporal artifacts, showing potential as a reward function for post-training video generation models.
\end{abstract}

\section{Introduction}
\label{intro}

Video quality assessment (VQA) seeks to predict perceived video quality in agreement with human perceptual judgments. 
% Early work targeted single distortions via designing expert models for specific degradations, such as blur~\citep{zhang2015perception,hassen2013image}, noise~\citep{golestaneh2013no,wang2002no}, and compression artifacts~\citep{golestaneh2013no,wang2002no}.
% Learning-based models have seen a major breakthrough via learning a mapping from scene-statistics-based features to human mean opinion scores (MOS)~\citep{cover2024cpvrws,wu2023dover,maxvqa,wang2021rich,wen2024modular, kou2024subjective}. 
The developments of large-scale video quality datasets such as 
LIVE-VQC~\citep{sinno2018large1,sinno2018large2, LIVE-VQC1}, KonViD-1k~\citep{konvid1k,hosu2017konstanz}, YouTube-UGC~\citep{wang2019youtube}, LSVQ~\citep{ying2021patch,lsvq} that provide diverse, realistic video content and reliable MOS annotations covering a wide range of content distributions, have fueled the advancements of end-to-end deep learning VQA models based on ConvNets~\citep{he2016deep,xie2017aggregated,simonyan2014two} or Transformers~\citep{yuan2021tokens, liu2021swin,tu2022maxvit} that directly learn to predict visual quality from raw pixels. 
Advanced neural architectures have been explored, such as the spatial-temporal attention~\citep{kim2018deep,martinez2019no,chen2021learning}, multi-scale feature fusion~\citep{xu2021perceptual,chen2021learning}, and spatially-sparse attention~\citep{wu2023discovqa,wu2022fast,wu2023neighbourhood} to capture complex spatiotemporal quality aspects.

However, a growing diversity of internet video content, from studio-made to user-generated content (UGC), to what is being predicted as the next wave of AI-generated content (AIGC), makes building a robust, generalizable VQA model increasingly challenging and costly. 
\underline{Firstly}, they often struggle to generalize under significant distribution shifts; for example, a VQA model trained on UGC has been found to suffer a substantial performance drop (around 70\% drop in PLCC for COVER~\cite{cover2024cpvrws} from UGC to AIGC) when evaluated on AIGC datasets with different distortion patterns.
\underline{Secondly}, existing models trained in an end-to-end manner lack \emph{interpretability}, although some works have explored interpretable approaches to some extent~\citep{wu2023dover,cover2024cpvrws,maxvqa}. This lack of diagnostic capability not only limits operational troubleshooting but also undermines trust in automated assessments.
\underline{Lastly}, the tightly coupled nature of most end-to-end designs makes extensibility cumbersome: adding a new feature, incorporating a novel distortion type, or adapting to emerging content classes often requires costly retraining or architectural redesign, making maintenance unscalable.

% [par3]

To address these challenges, we propose \textbf{\rvqa{}}, the first-of-its-kind agentic VQA system with expert routing and artifact localization for generic video content. 
Specifically, we leverage a vision-language models (VLMs) to analyze the video content, then reason over a routing pool of six state-of-the-art VQA expert models to generate a structured routing plan. Based on this plan, selected expert models are invoked to produce quality predictions, and then fused based on weights specified by the routing strategy by VLM. Together, these results are compiled into a comprehensive quality report that includes aggregated scores, per-expert breakdowns, and interpretable visual evidence to produce either a final quality score for standard VQA task or an answer for the VQA visual questioning task, offering both quantitative evaluation and actionable diagnostic insights. \rvqa{} offers three key advantages: (i) adaptability, by dynamically selecting relevant experts according to content and distortion characteristics; (ii) interpretability, by producing explicit routing rationales and diagnostic outputs; and (iii) robustness, by ensuring reliable performance across heterogeneous video domains, including both UGC and AIGC.
% [I think here you can focus on Tier 1] Then we will do xxxxxx tool-use xxxx.
% [The benefits are (i)(ii)(iii)]

% [par4]

To account for various inference budgets, we introduce three different tiers for \rvqa{}:
\ding{182} \textit{Tier 0}: a lightweight baseline that invokes a single expert model from the routing pool, providing fast yet coarse quality assessment;
\ding{183} \textit{Tier 1}: the standard configuration that the VLM performs coarse-grained reasoning and then assigns adaptive weights to experts accordingly;
\ding{184} \textit{Tier 2}: the full spatiotemporal artifact localization pipeline, which fuses expert predictions into a robust quality score while simultaneously identifying frame regions responsible for distortions, providing interpretable diagnostic evidence. The multi-tiered design enables \rvqa{} to support tasks from lightweight benchmarking to in-depth diagnostic analysis. Extensive experimental results demonstrate that \rvqa{} consistently achieves state-of-the-art performance across both standard VQA benchmarks and quality-related visual question answering tasks, validating its effectiveness, robustness, and versatility in diverse evaluation settings.

Our contributions are summarized as follows:

\begin{itemize}[leftmargin=*, parsep=0pt,topsep=0pt,itemsep=0pt]
    \item We present \rvqa{}, a first-of-a-kind agentic VQA framework that leverages a vision-language model to conduct structured reasoning and dynamic expert routing.

    \item We design a multi-tier system for \rvqa{} that adapts to different inference budgets, supporting tasks from lightweight benchmarking to in-depth diagnostic analysis.

    \item We introduce a spatial-temporal artifact localization pipeline that pinpoints distortions, offering novel interpretability and critical diagnostic evidence.
    
    \item We conduct comprehensive experiments on both UGC and AIGC video domains, showing that \rvqa{} consistently outperforms previous state-of-the-arts in terms of accuracy, robustness, and interpretability, across both video quality assessment and visual question answering tasks.  
\end{itemize}

\section{Methodology}
\label{method}

\begin{figure}[!t]
\centering
    \includegraphics[width=1.0\textwidth]{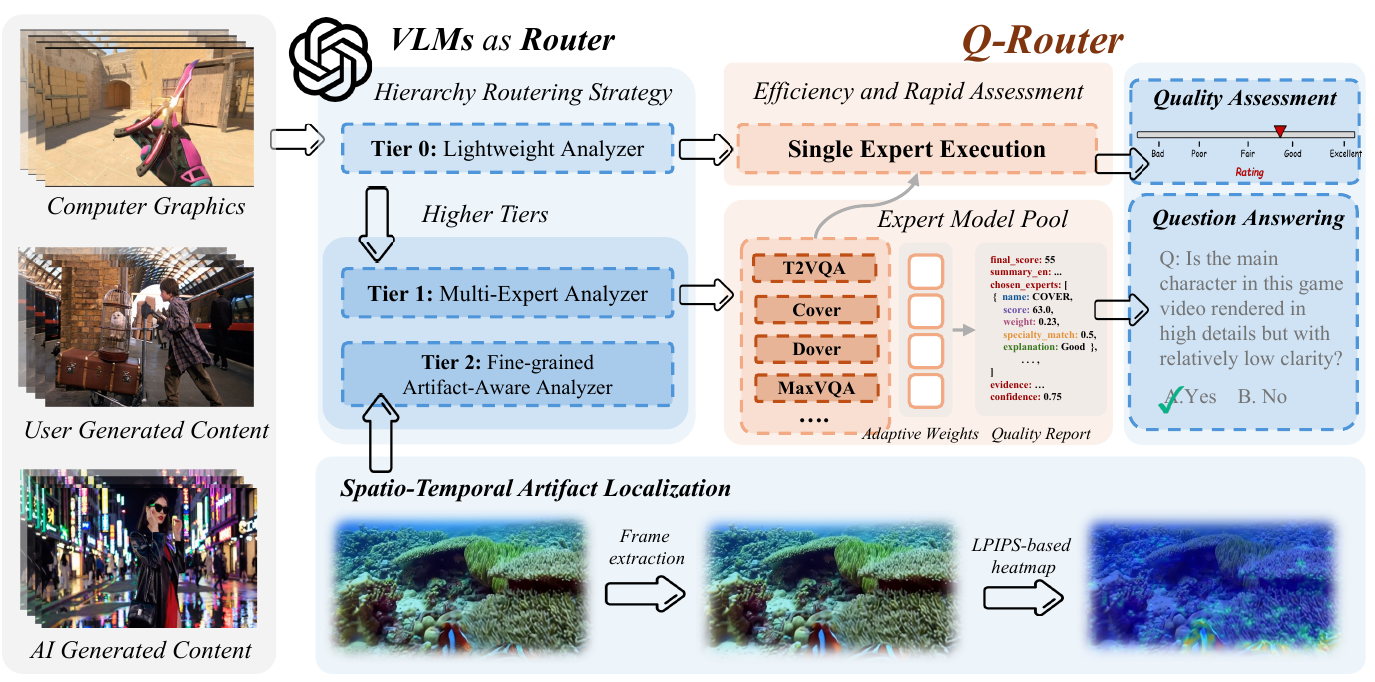}
    \captionof{figure}{We present \rvqa{},  an agentic framework designed for diverse video quality assessment tasks. \rvqa{} leverages a VLM as the router to dynamically assign the most suitable expert model from a comprehensive pool of state-of-the-art VQA methods. The expert pool includes COVER~\citep{cover2024cpvrws}, DOVER~\citep{wu2023dover}, BVQA~\citep{wen2024modular}, UVQA~\citep{wang2021rich}, MaxVQA~\citep{maxvqa}, and T2VQA~\citep{kou2024subjective}, enabling robust and adaptive evaluation across user-generated, AI-generated, and computer-generated video content.}
    \label{fig:teaser}
    \vspace{-3mm}
\end{figure}

In this section, we introduce \rvqa{}, an agentic routing framework designed for diverse video quality assessment (VQA) tasks. As shown in Figure~\ref{fig:teaser}, the \rvqa{} system employs a three-tiered routing hierarchy that organizes routing strategy and fusion operator according to varying cost–performance trade-off requirements. In practice, this design enables the user to dynamically adjust the inference pipeline based on available computational resources and the visual complexity of the input video. Specifically, \textbf{Tier 0} prioritizes efficiency and rapid assessment, while higher tiers (\textbf{Tier 1 \& 2}) leverage more specialized experts and richer fusion mechanisms to achieve greater accuracy and interpretability. The detailed design of these three tiers is outlined below.

% This hierarchical structure ensures that \rvqa{} can flexibly deploy to diverse scenarios, ranging from resource-constrained real-time applications to high-fidelity offline evaluations.

% 1. illustration router
% 2. pipeline of artifact localization
% output: rating, qa, report, heatmap 
% teaser: radar + ouput of the framework 

% \begin{wrapfigure}{r}{0.65\textwidth} % {r} = right side, width = 40% of text width
%     \centering
%     \vspace{-0.2in}
%     \includegraphics[width=\linewidth]{pics/qrouter-tasks.png}
%     \vspace{-0.2in}
%     \caption{The output of \rvqa{} can be flexibly adapted to a wide range of VQA tasks, enabling reliable quality evaluation across diverse video domains and application scenarios.}
%     \label{fig:wash-out_scaling}
%     \vspace{-5mm}
% \end{wrapfigure}

\paragraph{Tier 0} At the base level of \rvqa{}, the router operates in its most lightweight configuration by selecting a \textit{single} expert model from the routing pool based on the characteristics of the input video, such as its structural complexity, content modality, and observable quality attributes. Once the most appropriate expert is identified, the fusion operator directly adopts the prediction of this model as the \rvqa{}’s final output. By design, this tier prioritizes computational efficiency and fast inference, making it particularly suitable for scenarios where resource constraints or real-time processing requirements outweigh the need for more complex multi-expert aggregation.

\paragraph{Tier 1} At the intermediate tier, \rvqa{} expands the inference pipeline by enabling the router to select \textit{multiple} expert models rather than relying on a single candidate. The predictions from these experts are subsequently integrated through a relevance-aware weighted fusion operator, which dynamically adjusts weights according to the input’s characteristics. This design enhances robustness by leveraging complementary judgments across diverse experts, thereby improving both stability and predictive accuracy compared to \textbf{Tier~0}.  

\begin{figure}[ht]
    \centering
    % colback=gray!5!white,colframe=black!50!gray
    \begin{boxK}[colback=blue!5!white, colframe=blue!75!black, fontupper=\ttfamily\tiny]
    % , fontupper=\ttfamily\tiny
    
    You are VQA-Orchestrator, an expert-level agent for fusing multiple Video Quality Assessment (VQA) experts into a final quality score.
    \\
    \\
    \textbf{**Your task**}:
    \begin{enumerate}[leftmargin=2em]
        \item Given the video's metadata (type, description, orientation, etc.) and expert scores (all already scaled to [0-100]), dynamically assign weights to each expert based on their known biases, the video context, inter-model agreement, and confidence priors.

        \item Use these weighted scores to compute a **final\_score**:
        \begin{itemize}
            \item If `max(score) - min(score) > 20`, use weighted median; otherwise use weighted average.

            \item Round to the nearest integer in [0-100].
        \end{itemize}

        \item Produce a strictly formatted JSON output with:

        \begin{itemize}[leftmargin=2em]
            \item `final\_score`
            \item `summary\_en` ($\leq$120 characters English concise explanation including key factors and evidence)
            \item `chosen\_experts`, `per\_model` breakdown (score, weight, specialty match, notes)
            \item `evidence` (keyframe references, detected issues like banding/freeze, MaxVQA factors)
            \item `diagnostics` (score range, fusion method used, routing reasons, suggested next actions)
            \item `confidence` $\in$ [0,1]
        \end{itemize}
    \end{enumerate}

    \textbf{**Expert model cards (for routing logic)**:}
    \begin{itemize}[leftmargin=2em]
        \item **COVER**: Uses three parallel branches—technical (Swin Transformer), aesthetic (ConvNet), semantic (CLIP encoder)—combined via a cross-gating block to capture compression artifacts, aesthetic composition, and semantic coherence (real-time, ~96 fps)  \href{https://openaccess.thecvf.com/content/CVPR2024W/AI4Streaming/papers/He_COVER_A_Comprehensive_Video_Quality_Evaluator_CVPRW_2024_paper.pdf}{\textcolor{blue}{CVF Open Access}}.
        
        \item **DOVER++**: Extension of DOVER family; overlaps with COVER's aesthetic/technical branches; use primarily as a consistency reference unless its output aligns much better with most other experts.

        \item **UVQ**: Google's YouTube-trained universal VQA model built from millions of UGC examples; robust baseline when domain unclear or disagreements are large  \href{https://www.researchgate.net/publication/384424919_COVER_A_Comprehensive_Video_Quality_Evaluator}{\textcolor{blue}{ResearchGate}}] \href{https://research.google/blog/uvq-measuring-youtubes-perceptual-video-quality/?utm_source}{\textcolor{blue}{Google Research}}.

        \item MaxVQA (ExplainableVQA)**: CLIP-based, language-prompted model that provides both overall quality and fine-grained human-readable factors (e.g. banding, blur, aesthetic issues); used only for explanation and weight hints, not scoring \href{https://arxiv.org/abs/2305.12726?utm_source=chatgpt.com}{\textcolor{blue}{arXiv}}.

        \item **ModularBVQA**: Lightweight, modular baseline model suitable for low-latency deployment and serving as a fallback; modest sensitivity to capture-induced distortions.

        \item **T2VQA**: Text-to-video alignment model designed for AI-generated content, assessing fidelity between textual prompt and video; upweighted when video is AI-generated and textual condition is provided.
    \end{itemize}

    \textbf{**Video type → baseline weight priors**:}

    \begin{itemize}[leftmargin=2em]
        \item UGC: UVQ 0.25, COVER 0.25, ModularBVQA 0.15, RQ-VQA 0.10, MaxVQA 0.15  
        \item Short-form/social: RQ-VQA 0.30, COVER 0.30, UVQ 0.20, Modular 0.10, MaxVQA 0.10  
        \item Gaming: COVER-Technical 0.35, UVQ 0.25, Modular 0.20, MaxVQA 0.10, RQ-VQA 0.05  
        \item AI-Generated: T2VQA 0.35, COVER 0.20, UVQ 0.15, MaxVQA 0.15, Modular 0.10, RQ-VQA 0.05
    \end{itemize}

    \textbf{**Weight adjustment formula**:}
    weight\_i = base\_i x (1 + 0.5 x specialty\_match + 0.3 x agreement\_boost + 0.2 x confidence\_prior - 0.3 x oob\_penalty)
    Normalize weights to sum to 1. `specialty\_match` $\in \{1,0.5,0\}$ based on video type match. `agreement\_boost` = closeness to trimmed-mean of scores. `oob\_penalty` = penalize experts insensitive to detected issues or mismatched type.

    \end{boxK}

    \caption{Prompt for VQA with \rvqa{} (Tier 1) using GPT-4o.}
    \label{fig:tier1_prompt}
\end{figure}

\paragraph{Tier 2} At the advanced tier, \rvqa{} augments the inference pipeline by explicitly incorporating artifact localization into the fusion process. In this configuration, the router selects multiple expert models, while the fusion operator integrates spatially localized artifact maps that identify regions of potential quality degradation within frames or clips exhibiting artifacts. By leveraging these fine-grained signals, the system guides the fusion process toward distortion-prone areas, thereby enhancing both the accuracy and interpretability of the final assessment. This design not only strengthens robustness in challenging scenarios—such as AI-generated or heavily compressed content—but also grounds predictions in spatial evidence of distortions. By combining expert diversity with region-aware fusion, \textbf{Tier~2} delivers the highest level of reliability and diagnostic value within the \rvqa{} hierarchy. Moreover, the produced artifact maps can provide actionable insights supporting the post-processing tasks, such as targeted video restoration.

\subsection*{Spatio-temporal Artifact Localization}

In this section, we detail the process of conducting spatio-temporal artifact localization for Tier~2 of \rvqa{} (more details can be found in Appendix \ref{app:artifact-local}). Unlike Tier~0 and Tier~1, which operate solely on expert routing and aggregated scores, Tier~2 introduces fine-grained artifact localization across both spatial and temporal dimensions. This addition not only enhances the accuracy of video quality assessment but also provides interpretable evidence of model predictions.  

\paragraph{Probabilistic Frame Extraction}

Videos are first processed by the probabilistic extractor, which adaptively samples frames most likely to contain perceptual artifacts. Using low-level perceptual features (motion residuals, sharpness energy, gradient kurtosis, edge density, color histogram shifts, and optional semantic priors), each frame is assigned an artifact probability through a weighted logistic model. High-probability frames are grouped into temporally coherent clips via hysteresis thresholding and further refined by diversity sampling to ensure comprehensive coverage while remaining computationally efficient.

\begin{figure*}[htbp]
    \centering
    \begin{subfigure}{\textwidth}
    \includegraphics[height=3cm]{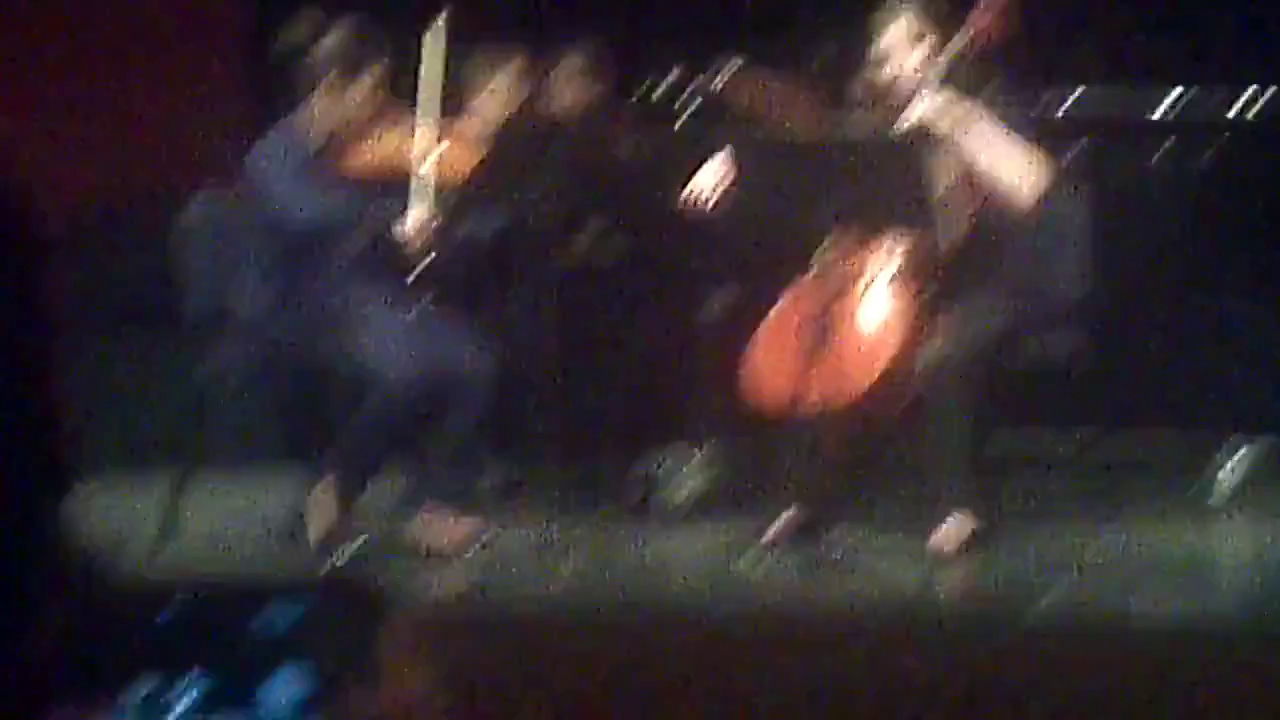}
    \includegraphics[height=3cm]{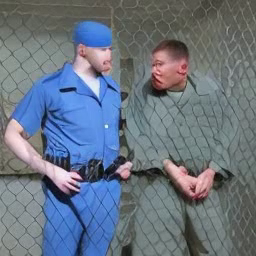}
    \includegraphics[height=3cm]{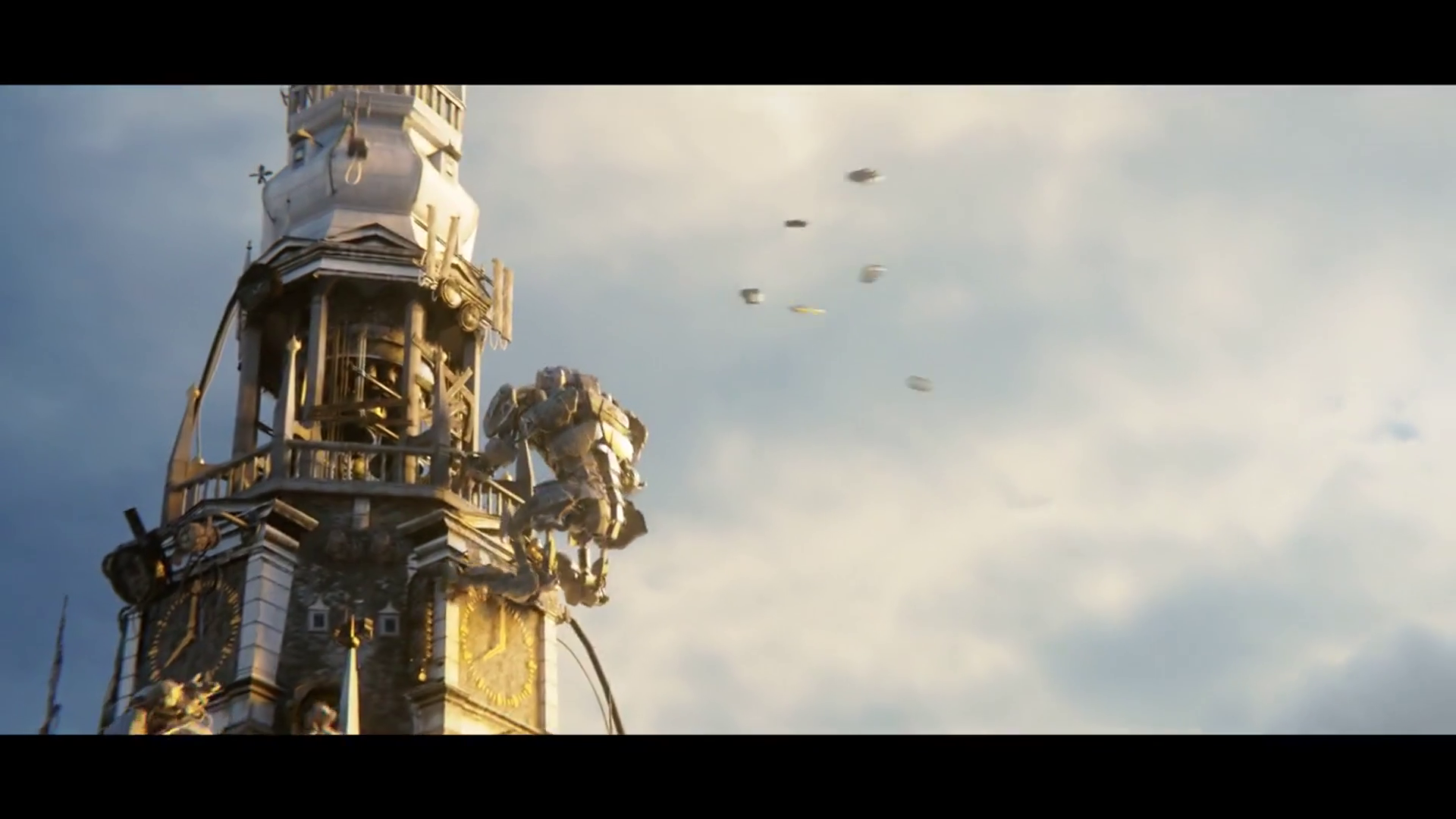}
    \subcaption{Examples of distorted frames from UGC, AIGC, and CG videos.}
    \includegraphics[height=3cm]{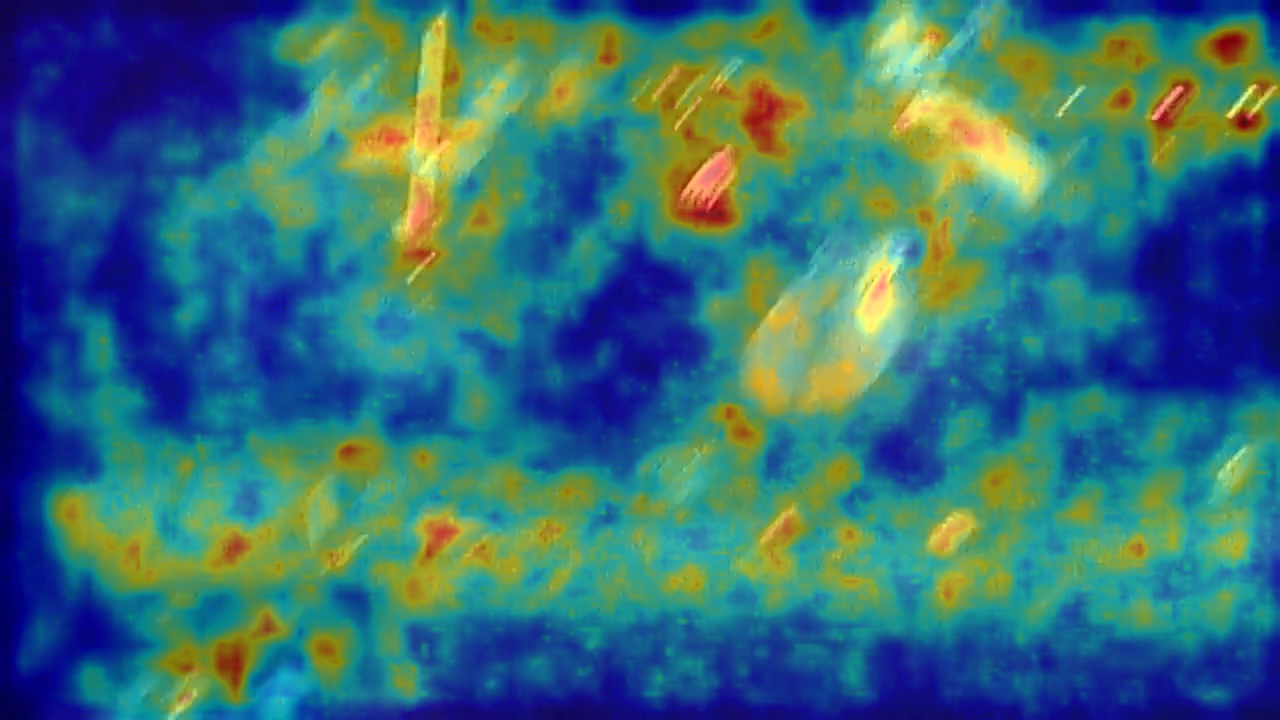}
    \includegraphics[height=3cm]{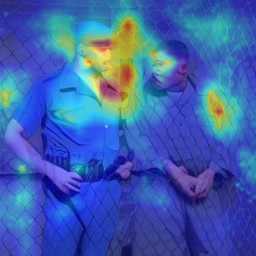}
    \includegraphics[height=3cm]{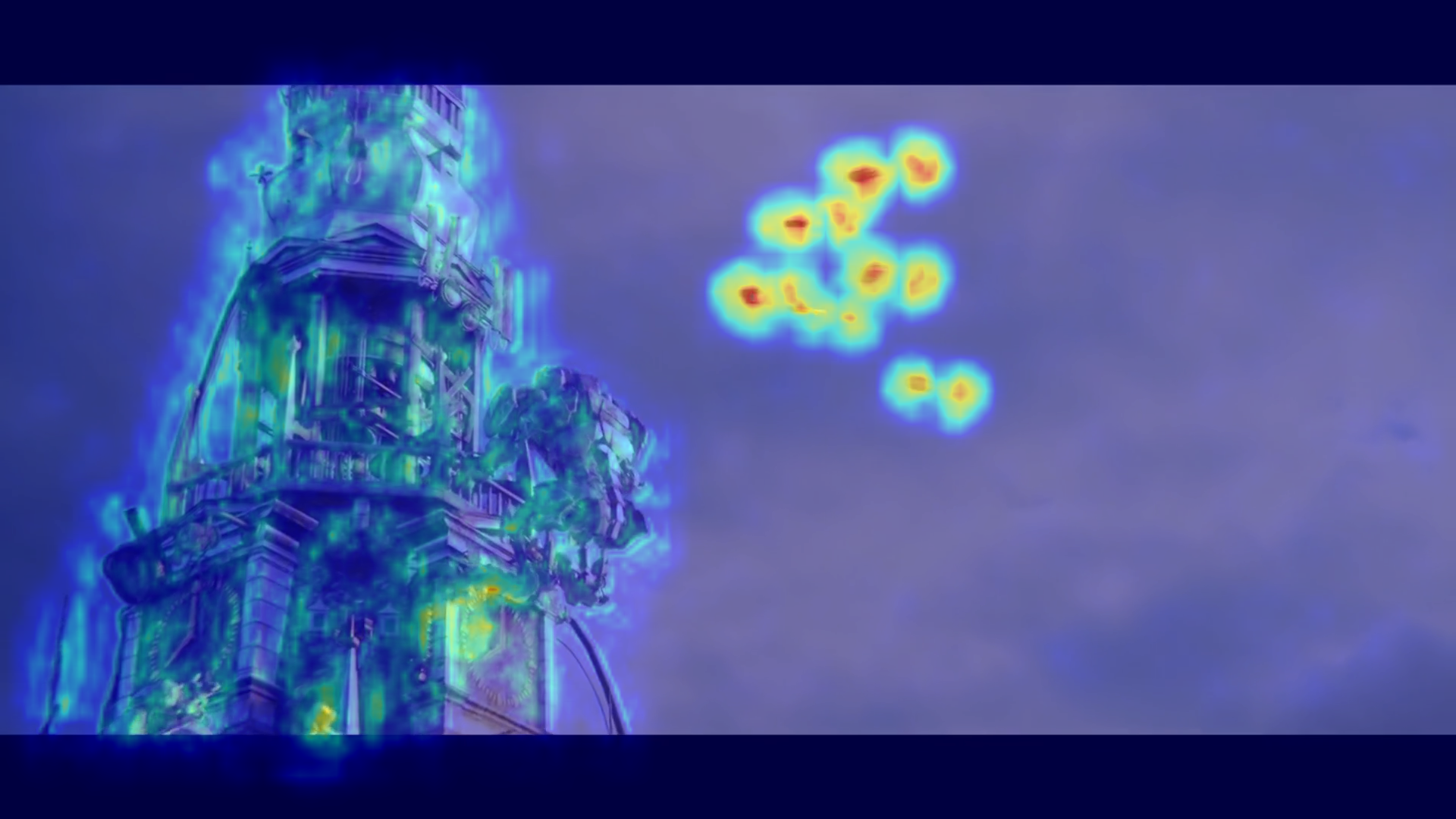}
    \subcaption{Corresponding artifact localization heatmaps for the distorted frames.}
    \label{fig:vis-b}   
    \end{subfigure}     
\caption{Distorted frames and their corresponding artifact localization heatmaps across UGC, AIGC, and CG videos. The first row shows example distorted frames, while the second row highlights suspicious regions detected by \rvqa{}.}\label{fig:visual} 
% \ashirbad{Reference Figure 2 in text}
\end{figure*}

\paragraph{VLM-Based Artifact Filtering}
To further improve the reliability of artifact detection, candidate frames from Step 1 are filtered using a GPT-4o model. Each frame is categorized into one of three artifact classes: (1) visual hallucinations (e.g., implausible objects or persons), (2) image artifacts (e.g., compression distortions, blurring, pixelation), or (3) AI-generated inconsistencies (e.g., unrealistic lighting or anatomy). Only frames flagged as containing artifacts are retained for localization, while clean frames are excluded.

\paragraph{Perceptual Heatmap Generation and Severity Estimation}
Within each retained clip, consecutive frame pairs are compared using the LPIPS~\citep{lpips} metric. To compensate for temporal motion, the subsequent frame is first aligned to the reference frame via optical flow (Farnebäck). The resulting LPIPS activation maps capture fine-grained perceptual differences and are normalized into interpretable heatmaps, as illustrated in Figure \ref{fig:visual}. Clip-level severity is then quantified by aggregating heatmap intensities via the mean pooling strategy, and the frame pair with the highest severity score is selected as the representative artifact instance for that clip. Finally, the corresponding heat map of the representative pair for each clip would be provided as an in-context augmentation for \rvqa{} during the evaluation of video quality.

\section{Experiments}
\label{exp}
We conduct two categories of experiments, corresponding to distinct VQA scenarios—quality rating and question answering—to validate the effectiveness of the proposed method. First, we evaluate \rvqa{} on standard VQA scoring datasets containing both UGC and AIGC videos. The results show that our method achieves consistently robust performance across benchmarks, outperforming baselines and achieving the best average results overall. Next, we assess the effectiveness of \rvqa{} on video quality question-answering tasks, with a focus on the Q-Bench-Video~\citep{qbenchvideo} benchmark. On this benchmark, our approach achieves a new state of the art, \textit{ranking at the top of the public leaderboard}.

\subsection{Video Quality Assessment}
\label{sec:vqa}

In this section, we evaluate \rvqa{} on VQA tasks where the objective is to rate the perceptual quality of input videos. This setting reflects the classical formulation of VQA, in which baseline models are pre-trained to predict quality scores that align with human subjective judgments.

\paragraph{Datasets} We conduct experiments on both UGC and AIGC VQA tasks. Specifically, for UGC, we evaluate our proposed method on widely used benchmarks including KoNViD-1k~\citep{konvid1k,hosu2017konstanz}, LSVQ-1080~\citep{ying2021patch,lsvq}, LSVQ-Test~\citep{ying2021patch,lsvq}, LIVE-VQC~\citep{sinno2018large1, sinno2018large2, LIVE-VQC1}, and Youtube-UGC (YT-UGC)~\citep{wang2019youtube}. For AIGC, we assess performance on the T2VQA-DB~\citep{kou2024subjective} dataset, which focuses on text-to-video generated content and poses unique challenges distinct from conventional UGC benchmarks.

\paragraph{Baselines} We benchmark \rvqa{} against several state-of-the-art models for UGC VQA, including COVER~\citep{cover2024cpvrws}, DOVER~\citep{wu2023dover}, MaxVQA~\cite{maxvqa}, ModularBVQA~\citep{wen2024modular}, and UVQ~\citep{wang2021rich}, which represent diverse design paradigms for no-reference VQA. For AIGC VQA, we adopt T2VQA~\citep{kou2024subjective} as the baseline, which is developed on the T2VQA-DB dataset specifically for evaluating text-to-video generated content.

\begin{table}[htbp]
\footnotesize
\setlength{\tabcolsep}{4.5pt}
\centering
\resizebox{\columnwidth}{!}{%

\begin{tabular}{l|l|ccccccc}
\toprule
\bf Dataset & \bf Metric & \bf COVER & \bf DOVER & \bf MaxVQA & \bf BVQA & \bf T2VQA & \bf UVQ & \bf \rvqa{} \\
\midrule
  \multicolumn{9}{c}{\textit{\cellcolor{gray!10} UGC Video Quality Assessment Benchmark}} \\
\midrule
\multirow{2}{*}{KoNViD-1k} 
  & PLCC & 0.8528 & 0.8447 & \underline{0.8627} & \textbf{0.8731} & 0.1717 & 0.6998 & 0.8612 \\
  & SRCC & 0.8507 & 0.8341 & 0.8599 & \textbf{0.8713} & 0.1510 & 0.7067 & \underline{0.8607} \\
\midrule
\multirow{2}{*}{LSVQ-1080} 
  & PLCC & \underline{0.7806} & 0.7811 & 0.7419 & 0.7337 & 0.2371 & 0.5304 & \textbf{0.8076} \\
  & SRCC & 0.7544 & \textbf{0.7808} & 0.7282 & 0.7431 & 0.2479 & 0.5068 & \underline{0.7792} \\
\midrule
\multirow{2}{*}{LSVQ-Test} 
  & PLCC & 0.8378 & \textbf{0.8660} & 0.8005 & \underline{0.8583} & 0.2368 & 0.6801 & 0.8540 \\
  & SRCC & 0.8465 & \textbf{0.8777} & 0.8100 & 0.8514 & 0.2460 & 0.6858 & \underline{0.8585} \\

\midrule
\multirow{2}{*}{LIVE-VQC} 
  & PLCC & 0.7624 & \underline{0.8742} & 0.7325 & 0.8571 & 0.1719 & 0.6827 & \textbf{0.8806} \\
  & SRCC & 0.7254 & \underline{0.8187} & 0.6942 & 0.8103 & 0.2015 & 0.6800 & \textbf{0.8293} \\
\midrule
\multirow{2}{*}{YT-UGC} 
  & PLCC & 0.1529  & 0.7367  &  0.7761 &  0.6974 &  0.2901 &  \textbf{0.9402} & \underline{0.8920}  \\
  & SRCC &  0.1413 & 0.7195  & 0.7789  &  0.6749 & 0.2777  &  \textbf{0.9457} &  \underline{0.8909} \\

  \midrule
  \multicolumn{9}{c}{\textit{\cellcolor{gray!10}  AIGC Video Quality Assessment Benchmark}} \\
    \midrule
  \multirow{2}{*}{T2VQA-DB} 
  & PLCC & 0.2468 & 0.0642 & 0.2361 & 0.1051 & \underline{0.7733} & 0.1158 & \textbf{0.8283} \\
  & SRCC & 0.1377 & 0.0429 & 0.2010 & 0.0691 & \underline{0.7828} & 0.0983 & \textbf{0.8258} \\
  \midrule
  \multicolumn{9}{c}{\textit{\cellcolor{gray!10} Overall Performance}} \\
\midrule
\multirow{2}{*}{Average} 
  & PLCC & 0.6056  & \underline{0.6945}  &  0.6916 & 0.6874 &  0.3135 & 0.6081  &  \textbf{0.8539} \\
  & SRCC & 0.5760  & \underline{0.6789}  & 0.6787  &  0.6700 &  0.3178 &  0.6038 &  \textbf{0.8407} \\
\bottomrule
\end{tabular}
}
\caption{Comparison of methods on UGC and AIGC benchmarks against the \rvqa{} (Tier 1). Best results are highlighted in bold, while the second-best results are underlined.}
\label{tab:vqa-results}
\end{table}

\paragraph{Experimental Setup} We employ GPT-4o~\citep{gpt4o, openai2024gpt4technicalreport} as the backbone router and fusion operator within \rvqa{}. The router is responsible for analyzing the input video descriptors and selecting appropriate expert models, while the fusion operator integrates their outputs into a unified quality report and produces the final quality score on a $[0,100]$ scale. For reproducibility, the complete prompts used to deploy GPT-4o, including routing instructions and fusion templates, are provided in Appendix~\ref{app:prompts}.

As for the routing pool, we include all the baseline models evaluated in this section, which collectively serve as the candidate experts. For UGC VQA, we leverage the publicly released model weights to ensure faithful implementation and evaluation. Since T2VQA does not provide open-sourced model weights, we reproduce its training process by following the official repository\footnote{https://github.com/QMME/T2VQA}.

\paragraph{Results}

Table~\ref{tab:vqa-results} summarizes the performance of \rvqa{} compared with state-of-the-art baselines on both UGC and AIGC VQA benchmarks. Our method achieves the best average performance across datasets, with a PLCC of $0.8539$ and SRCC of $.8407$, outperforming all individual baselines by a significant margin over $20\%$, highlighting the effectiveness of the routing-and-fusion paradigm in leveraging complementary strengths of heterogeneous expert models.

On UGC datasets, \rvqa{} consistently demonstrates superior or highly competitive results. For example, it achieves the best PLCC ($0.8076$) on LSVQ-1080 and the second-best SRCC ($0.8607$) on KoNViD-1k, surpassing strong baselines such as COVER~\citep{cover2024cpvrws}, DOVER~\citep{wu2023dover}, and MaxVQA~\citep{maxvqa}. Importantly, \rvqa{} maintains robustness across different datasets, whereas individual baselines often exhibit dataset-specific strengths but lack generalization.

On the T2VQA-DB dataset, \rvqa{} significantly outperforms all baselines, reaching a PLCC of $0.8283$ and SRCC of $0.8258$. This demonstrates the \rvqa{}’s ability to adaptively leverage expert knowledge and handle distribution shifts in AI-generated video quality—a setting where conventional UGC-trained models fail (PLCC $< 0.25$).

\subsection{Visual Question Answering on Video Quality}
\label{subsec:vqa-exp}

\paragraph{Datasets} We further conduct experiments on visual question answering tasks related to video quality. Specifically, we evaluate a wide range of VLMs, along with \rvqa{}, on Q-Bench-Video~\citep{qbenchvideo}, which is the first and currently only benchmark dedicated to this task.

\paragraph{Baselines}
We deploy \rvqa{} with GPT-4o~\citep{gpt4o,openai2024gpt4technicalreport} as the backbone, and evaluate it alongside several widely used open-sourced VLMs, including LLaVA-Next~\citep{llavanext}, LLaVA-v1.5~\citep{liu2024llava}, mPLUG-Owl2~\citep{ye2024mplug2}, mPLUG-Owl3~\citep{ye2024mplug3}, LLaVA-OneVision~\citep{zhang2025llava}, InternVL-Chat~\citep{chen2024far}, VILA1.5~\citep{ke2023vila}, PLLaVA~\citep{xu2024pllava}, LLaVA-Next-Video~\citep{llavanext}, ST-LLM~\citep{liu2024st}, Video-LLaVA~\citep{lin2023video}, VideoChat2~\citep{li2023videochat}, and GPT-4o~\citep{gpt4o}. These models represent the current generation of multimodal models designed for video understanding and serve as competitive baselines. In addition, we include GPT-4o itself as a strong proprietary baseline, as it currently ranks as the state-of-the-art model on the Q-Bench-Video leaderboard~\citep{qbenchvideo}.

\paragraph{Experimental Setup} We adopt the same experimental settings as described in Section~\ref{sec:vqa} for video quality rating, ensuring consistency across evaluation protocols. Specifically, we employ GPT-4o~\citep{gpt4o} as the backbone router and fusion operator within \rvqa{}, and construct the routing pool using the same set of expert models introduced earlier. The evaluation is conducted on Q-Bench-Video~\citep{qbenchvideo}, a large-scale benchmark designed for video quality question answering, which covers diverse question types (e.g., Yes-or-No, What/How), video modalities (UGC, AIGC, and CG), and contextual reasoning settings (e.g., Referring, Global, and Comparison). To ensure reproducibility, we follow the official evaluation protocol of Q-Bench-Video and report accuracy across all categories as well as the overall score.

\begin{table*}[ht]
\footnotesize
\setlength{\tabcolsep}{3pt}
\centering
\resizebox{\columnwidth}{!}{%

\begin{tabular}{l|c|ccc|cccccccc}
\toprule
\multirow{2}{*}{\bf Method} & \multirow{2}{*}{\bf Overall} & \multicolumn{3}{c}{\bf Question type} & \multicolumn{4}{c}{\bf Quality Concern} \\
\cmidrule(lr){3-5} \cmidrule(lr){6-9}
 & & \bf Yes-or-No & \bf What/How & \bf Open-ended & \bf Technical & \bf Aesthetic & \bf Temporal & \bf AIGC \\
% \midrule
% Random G uess & 37.27 & 50.00 & - & 25.00 &  36.76 & 37.30 & 37.27 & 37.62 \\
\midrule
\makecell[l]{LLaVA-Next}  & 47.00 &63.20 & 43.78 & 30.42 & 45.95 & 54.83 & 45.63 & 46.24 \\
 \makecell[l]{LLaVA-v1.5}  & 45.57 & 53.40 & 46.87 & 33.85 & 55.83 & 55.90 & 44.91 & 45.96\\
\makecell[l]{mPLUG-Owl2} & 44.20 & 59.61 & 38.83 & 31.57 & 42.49 & 53.28 & 44.73 & 40.07 \\
 \makecell[l]{mPLUG-Owl3} & 52.44 & 60.82 & 56.52 & 35.84 & 51.34 & 60.46 & 54.26 & 37.30  \\
\makecell[l]{LLaVA-OneVision}  & 52.12 & 62.13 & 52.23 & 38.56 & 48.74 & 61.53 & 48.81 & 44.57 \\
\makecell[l]{InternVL-Chat} & 51.91  & 70.21 & 48.65 & 32.20 & 50.24 & 49.50 & 52.96  & 47.69\\
\makecell[l]{VILA1.5}  & 49.62 & 61.59 & 47.30 & 36.88 & 46.74 & 59.30 & 47.57 & 43.67 \\
\makecell[l]{PLLaVA}  & 51.23 & 65.13 & 54.23 & 29.44 & 50.31 & 60.09 & 50.13 & \underline{50.75} \\
\makecell[l]{LLaVA-Next-Video} & 48.97 & 65.98 & 45.31 & 31.92 & 48.11 & 57.33 & 47.09 & 45.56  \\
\makecell[l]{ST-LLM} & 35.89 & 46.43 & 28.45 & 32.31 & 33.32 & 45.66 & 36.01 & 32.62 \\
\makecell[l]{Video-LLaVA}  & 45.89 & 64.36 & 39.38 & 30.86 & 43.35 & 55.97 & 45.58 & 43.64 \\
 \makecell[l]{VideoChat2}  & 42.06 & 56.54 & 33.13 & 35.36 & 39.09 & 49.27 & 41.59 & 38.04 \\
% InternVL-Chat & 51.11 & 66.02 & 52.13 & 48.42 & 52.73 & 50.59 & 53.12 \\
% LLaVA-Next-Video & 48.69 & 61.34 & 45.95 & 49.03 & 60.94 & 46.97 & 49.40 \\
% Video-LLaVA & 43.49 & 64.67 & 40.79 & 43.25 & 54.04 & 42.38 & 42.76 \\
% LLaVA-OneVision & 51.70 & 61.34 & \underline{53.88} & 49.35 & \textbf{64.15} & 50.68 & 44.30 \\
GPT-4o$^\dag$ & 56.91 & 69.95 & 55.20 & 42.10 & 55.74 & \textbf{66.06} & 57.66 & 44.25 \\
\midrule
\rowcolor[gray]{0.9}  \rvqa{} \textit{(Tier 0)} & 56.59 & 74.82 & 54.30 & 35.80 & 54.86 & 51.34 & 54.91 & \textbf{56.67}  \\
\rowcolor[gray]{0.9}  \rvqa{} \textit{(Tier 1)} & \underline{59.40} & \underline{75.53} & \underline{56.33} & \underline{42.44} & \textbf{59.48} & 62.90 & \underline{58.45} & 45.00 \\
\rowcolor[gray]{0.9}  \rvqa{} \textit{(Tier 2)} & \textbf{60.07} & \textbf{76.00} & \textbf{57.01} & \textbf{43.33} & \underline{59.31} & \underline{65.63} & \textbf{61.46} & 50.22 \\
\bottomrule
\end{tabular}
}
\caption{Comparison of methods on the closed-ended questions in Q-Bench-Video \texttt{dev} split across question types and quality concerns. The best results are highlighted in bold, while the second-best results are underlined. $^\dag$: As the models continue to evolve, we reproduce and report the results of the GPT-4o model for comparison.}
\label{tab:qbench-dev}
\end{table*}

\paragraph{Results}

Table~\ref{tab:qbench-dev} presents the performance of \rvqa{} on the Q-Bench-Video \texttt{dev} split, compared against a set of strong VLM baselines. Our proposed routing framework achieves consistent improvements across all tiers. Tier 0, which routes to a single expert, already matches GPT-4o (56.59 vs. 56.91 overall), while substantially outperforming it on AIGC-related questions (56.67 vs. 44.25). Tier 1, which leverages weighted expert fusion, further improves to 59.40 overall, with strong gains in technical (59.48) and aesthetic (62.90) dimensions. Finally, Tier 2, which integrates artifact localization into the fusion stage, delivers the best results overall (60.07), surpassing GPT-4o. Tier 2 achieves new state-of-the-art performance across most categories, notably Yes-or-No (76.00), What/How (57.01), open-ended (43.33), and temporal concerns (61.46).
\setlength{\tabcolsep}{4pt}
\renewcommand{\arraystretch}{0.95}

\subsection{Ablation Study}

In this section, we conduct the ablation study on the key component of our proposed \rvqa{} framework. Specifically, we examine the performance of \rvqa{} under different prompting structures:, including \textbf{(1)} \textit{Zero-shot Prompting}: the VLM is directly prompted to output the predicted rating for the input video on a scale of $[0,100]$; \textbf{(2)} \textit{CoT prompting}: The prompt template includes background information on the dimensions and aspects that should be considered when assessing the video quality; \textbf{(3)} \textit{Expert Model Fusing}: directly prompt the VLM to fuse the results from expert models.; \textbf{(4)} the comprehensive \textit{\rvqa{} Tier 1} framework. All experiments in this section are conducted using the \rvqa{} (Tier 1) configuration and GPT-4o as the backbone. All experiments use the Tier 1 configuration with GPT-4o as the backbone, following the same setup as Section~\ref{subsec:vqa-exp}.

\begin{figure*}[htbp]
    \centering
    \includegraphics[width=0.49\linewidth]{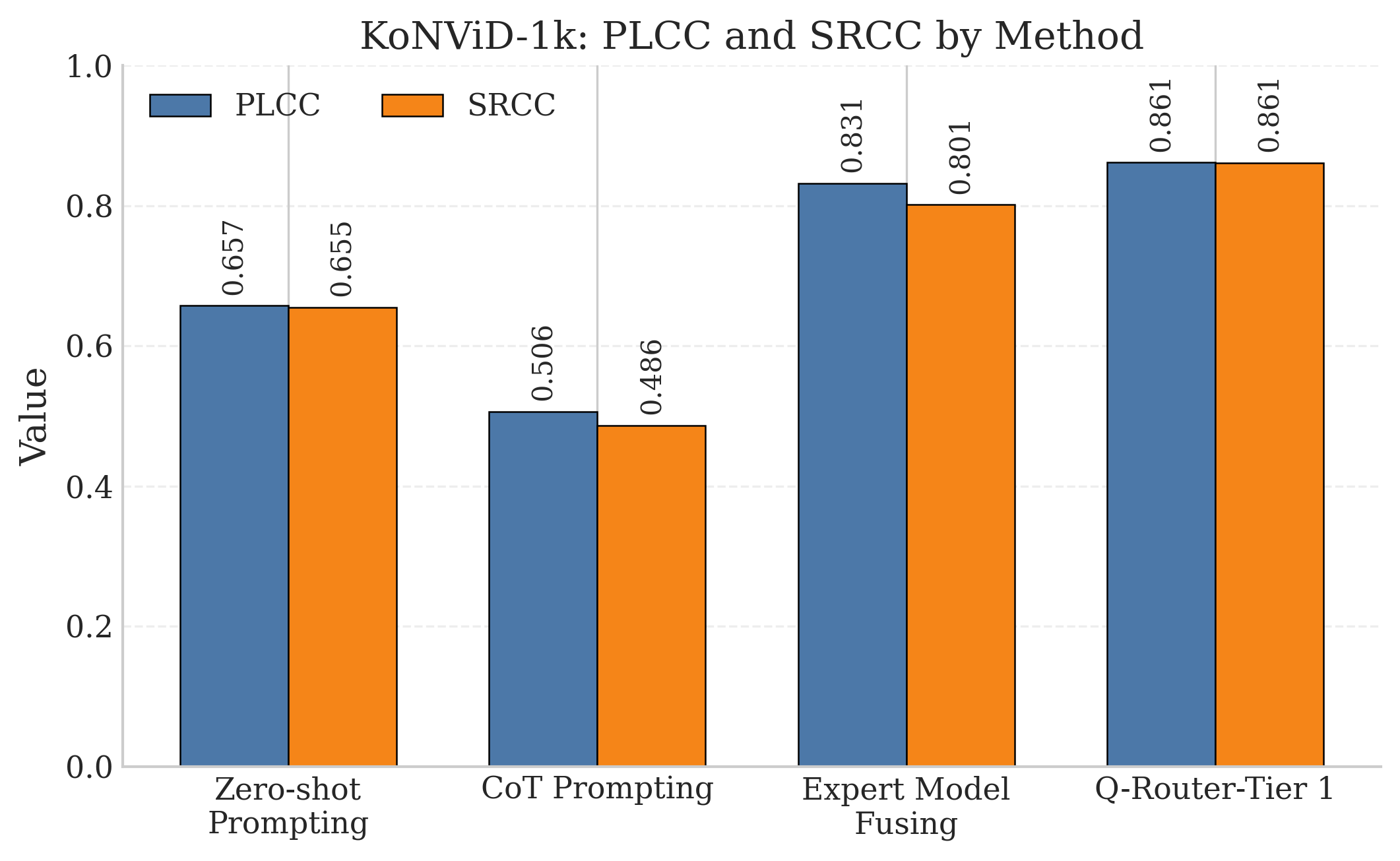}
    \includegraphics[width=0.49\linewidth]{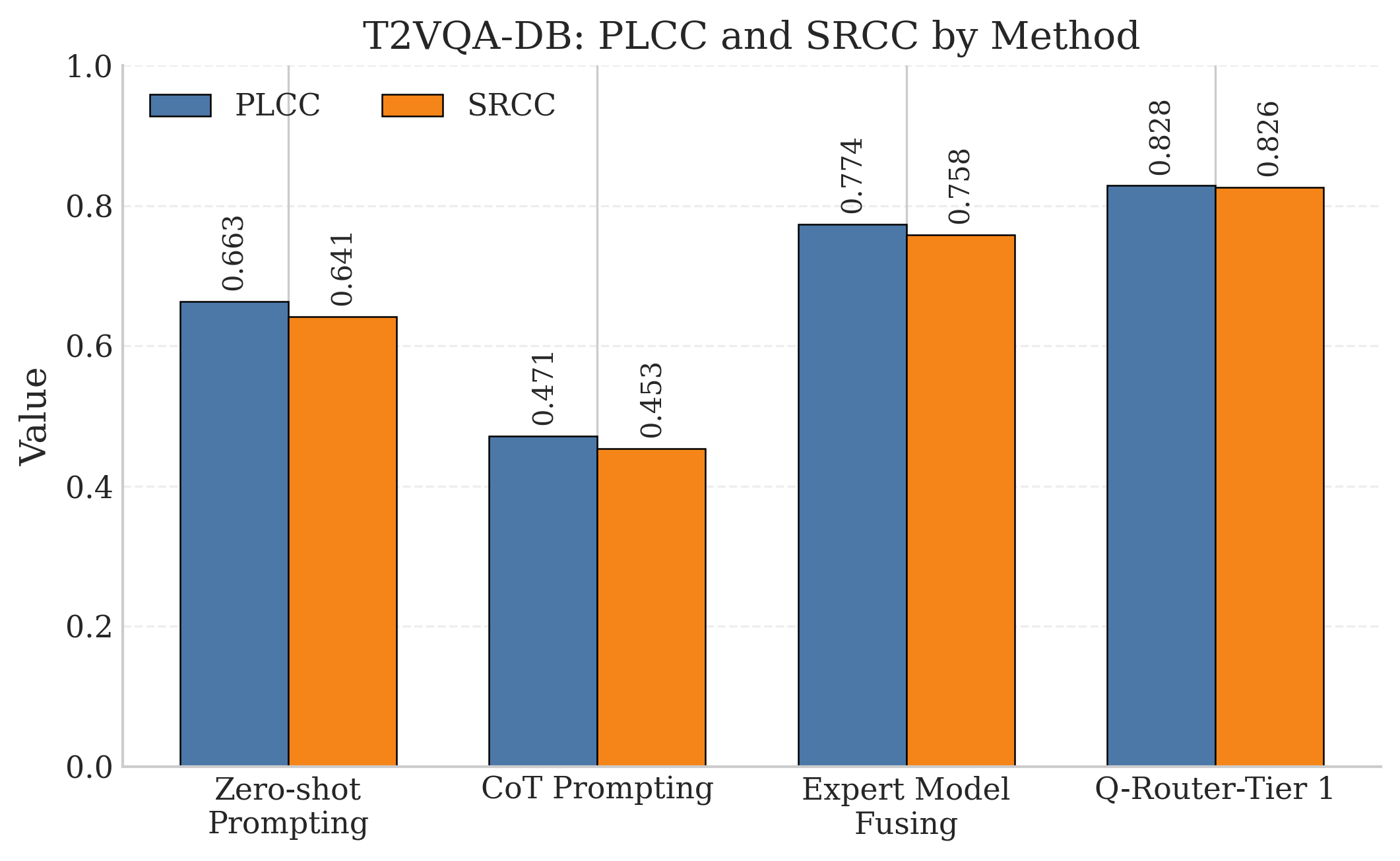}
    \label{fig:vis-b}   
\caption{Impact of prompting structure across both UGC and AIGC benchmarks for VQA, with GPT-4o as backbone.}\label{fig:abaltion} 
% \ashirbad{Reference Figure 2 in text}
\end{figure*}

As presented in Figure \ref{fig:abaltion}, zero-shot prompting results exhibit surprisingly high correlation with MOS, reflecting the effectiveness of VLM's system 1 thinking on VQA tasks. While CoT prompting performs substantially worse, suggesting that introducing forced deliberation through chain-of-thought may disrupt alignment with human perception in the VQA setting. Additionally, the expert model fusing strategy achieves a notable performance boost, confirming the complementary strengths of multiple specialized VQA models. Finally, our proposed \rvqa{} (Tier 1) yields the highest performance across both benchmarks, indicating the adative routing framework not only leverage the expert knowledge but also preserves alignment with human preference.

\section{Related Works}
\label{related_works}

\paragraph{Video Quality Assessment} Traditional approaches to video quality assessment (VQA) largely rely on full-reference (FR) and reduced-reference (RR) metrics, such as PSNR, SSIM~\citep{wang2004image}, and VMAF~\citep{Li_VMAF_2016}. To address the reliance on reference videos for FR, no-reference (NR) models were introduced, estimating quality directly from distorted videos by extracting handcrafted features related to motion, texture, and temporal consistency~\citep{saad2014blind, li2016spatiotemporal}. However, the generalization ability of handcrafted features remains limited, particularly for diverse real-world content. The advent of deep learning has substantially advanced NR-VQA. Early CNN-based models focused on learning spatial representations~\citep{kim2018deep}, while later works incorporated temporal modeling~\citep{chen2021deep,li2019quality}. Large-scale benchmarks such as KoNViD-1k~\citep{konvid1k}, LSVQ~\citep{ying2021patch}, and YouTube-UGC~\citep{wang2019youtube} have provided the foundation for training and evaluating deep VQA models, fostering rapid progress in the field. Recent works have explored modular and expert-driven designs for VQA. Methods such as COVER~\citep{cover2024cpvrws}, DOVER~\citep{wu2023dover}, and MaxVQA~\citep{maxvqa} leverage multi-branch architectures or ensembles to account for diverse perceptual factors. However, most existing approaches adopt fixed fusion strategies, limiting adaptivity across heterogeneous inputs~\cite{zheng2024video}. In contrast, our work introduces \rvqa{}, a routing framework that leverages VLMs to dynamically select and fuse expert predictions, offering a flexible and extensible solution that adapts to diverse video inputs while maintaining interpretability.

\paragraph{LLM Reasoning and Agent} 
Large Language Models (LLMs)~\citep{devlin2018bert, radford2019gpt2,brown2020gpt3,team2023gemini,roziere2023codellama,touvron2023llama,touvron2023llama2, raffel2020t5,qwen2,qwen2.5,pan2024plum} have demonstrated remarkable capabilities on complex reasoning and understanding tasks across natural language, code generation, and multimodal domains. Reasoning techniques/framework like Chain-of-Thought (CoT) prompting~\citep{wei2022chain}  and self-consistency~\citep{wang2022self} have been shown to significantly enhance reasoning performance by encouraging models to generate intermediate steps rather than producing direct answers. More recent works explore decompositional reasoning~\citep{zhou2022least} and tool-augmented reasoning~\citep{schick2023toolformer}, highlighting the ability of LLMs to integrate the structured process and external knowledge. 

Building on these advances, Large Vision Language Models (VLMs)~\citep{li2022blip, li2023blip2, liu2024llava,llavanext,llama3.2, Qwen-VL, Qwen2VL, lu2024deepseek, wu2024deepseek, bai2025qwen2} extended the understanding and reasoning capabilities of LLMs into the visual domain. Recent works demonstrate that VLMs can perform not only captioning or classification, but also higher-level reasoning over visual inputs in real-world application scenarios~\citep{moor2023med,li2024llava-med,shao2024lmdrive,tian2024drivevlm,sima2023drivelm,openemma, ma2025position,wang2025generative,rana2023sayplan,kim2024openvla, xing2025can}.

A growing line of work has begun to frame LLMs and VLMs as agents capable of orchestrating modular pipelines. Prior studies demonstrate that such models can act as routers, dynamically selecting and invoking specialized tools or experts to solve complex tasks~\citealp{schick2023toolformer, yao2023react, sumers2023cognitive, shen2023hugginggpt,luo2025large,zuo20254kagent}. These agentic approaches emphasize the potential of LLMs/VLMs not merely as standalone predictors but as controllers of adaptive, extensible systems—a paradigm shift that directly inspires our design of \rvqa{} as a routing-based framework for video quality assessment.

\section{Conclusion}
In this work, we introduced \rvqa{}, an agentic framework for video quality assessment that leverages VLMs as routers and fusion operators over a diverse pool of expert models. By structuring the inference process into a three-tier hierarchy—ranging from lightweight single-expert routing, to multi-expert weighted fusion, and finally to artifact-localized fusion—\rvqa{} achieves a balance between efficiency, robustness, and interpretability.

Extensive experiments on both classical VQA benchmarks (UGC and AIGC) and the Q-Bench-Video question-answering benchmark demonstrate that \rvqa{} consistently outperforms state-of-the-art baselines. Beyond improving performance, \rvqa{} enhances interpretability by grounding outputs in spatial evidence of distortions and offers actionable insights that can inform downstream tasks like restoration and adaptive post-processing. These results highlight the promise of expert routing as a paradigm for advancing video quality assessment. Furthermore, the modular nature of \rvqa{} provides opportunities for extending the framework to broader multimodal evaluation tasks, incorporating new expert models as they emerge, and exploring more advanced routing strategies that further improve efficiency and generalization.

% \newpage

% \section{Ethics Statement}
% Our work on \rvqa{} adheres to the ICLR Code of Ethics\footnote{https://iclr.cc/public/CodeOfEthics}
% . The research does not involve human subjects, personally identifiable information, or sensitive attributes, and therefore does not raise concerns related to human experimentation, privacy, or consent. All datasets used are publicly available under appropriate licenses, and no proprietary or confidential data were employed.

% \section{Reproducibility statement}
% We comprehensively presnet the implementation details in Sectio~\ref{subsec:practice}, Appendix ~\ref{app:artifact-local}, and \ref{app:prompts}. All the evaluation code are available from public online code repositories. All data and source code of \rvqa{} pipeline will be released upon acceptance if they are not from online codebase.

\bibliography{main}

\begin{thebibliography}{10}

\bibitem{sinno2018large1}
Zeina Sinno and Alan~Conrad Bovik.
\newblock Large-scale study of perceptual video quality.
\newblock {\em IEEE Transactions on Image Processing}, 28(2):612--627, 2018.

\bibitem{sinno2018large2}
Zeina Sinno and Alan~C Bovik.
\newblock Large scale subjective video quality study.
\newblock In {\em 2018 25th IEEE International Conference on Image Processing (ICIP)}, pages 276--280. IEEE, 2018.

\bibitem{LIVE-VQC1}
Zeina Sinno and Alan~Conrad Bovik.
\newblock Large-scale study of perceptual video quality.
\newblock {\em IEEE Transactions on Image Processing}, 28(2):612--627, 2019.

\bibitem{konvid1k}
Vlad Hosu, Franz Hahn, Mohsen Jenadeleh, Hanhe Lin, Hui Men, Tam{\'a}s Szir{\'a}nyi, Shujun Li, and Dietmar Saupe.
\newblock The konstanz natural video database, 2017.

\bibitem{hosu2017konstanz}
Vlad Hosu, Franz Hahn, Mohsen Jenadeleh, Hanhe Lin, Hui Men, Tam{\'a}s Szir{\'a}nyi, Shujun Li, and Dietmar Saupe.
\newblock The konstanz natural video database (konvid-1k).
\newblock In {\em 2017 Ninth International Conference on Quality of Multimedia Experience (QoMEX)}, pages 1--6. IEEE, 2017.

\bibitem{wang2019youtube}
Yilin Wang, Sasi Inguva, and Balu Adsumilli.
\newblock Youtube ugc dataset for video compression research.
\newblock In {\em 2019 IEEE 21st international workshop on multimedia signal processing (MMSP)}, pages 1--5. IEEE, 2019.

\bibitem{ying2021patch}
Zhenqiang Ying, Maniratnam Mandal, Deepti Ghadiyaram, and Alan Bovik.
\newblock Patch-vq:'patching up'the video quality problem.
\newblock In {\em Proceedings of the IEEE/CVF conference on computer vision and pattern recognition}, pages 14019--14029, 2021.

\bibitem{lsvq}
Zhenqiang Ying, Maniratnam Mandal, Deepti Ghadiyaram, and Alan Bovik.
\newblock Live large-scale social video quality (lsvq) database, 2021.

\bibitem{he2016deep}
Kaiming He, Xiangyu Zhang, Shaoqing Ren, and Jian Sun.
\newblock Deep residual learning for image recognition.
\newblock In {\em Proceedings of the IEEE conference on computer vision and pattern recognition}, pages 770--778, 2016.

\bibitem{xie2017aggregated}
Saining Xie, Ross Girshick, Piotr Doll{\'a}r, Zhuowen Tu, and Kaiming He.
\newblock Aggregated residual transformations for deep neural networks.
\newblock In {\em Proceedings of the IEEE conference on computer vision and pattern recognition}, pages 1492--1500, 2017.

\bibitem{simonyan2014two}
Karen Simonyan and Andrew Zisserman.
\newblock Two-stream convolutional networks for action recognition in videos.
\newblock {\em Advances in neural information processing systems}, 27, 2014.

\bibitem{yuan2021tokens}
Li~Yuan, Yunpeng Chen, Tao Wang, Weihao Yu, Yujun Shi, Zi-Hang Jiang, Francis~EH Tay, Jiashi Feng, and Shuicheng Yan.
\newblock Tokens-to-token vit: Training vision transformers from scratch on imagenet.
\newblock In {\em Proceedings of the IEEE/CVF international conference on computer vision}, pages 558--567, 2021.

\bibitem{liu2021swin}
Ze~Liu, Yutong Lin, Yue Cao, Han Hu, Yixuan Wei, Zheng Zhang, Stephen Lin, and Baining Guo.
\newblock Swin transformer: Hierarchical vision transformer using shifted windows.
\newblock In {\em Proceedings of the IEEE/CVF international conference on computer vision}, pages 10012--10022, 2021.

\bibitem{tu2022maxvit}
Zhengzhong Tu, Hossein Talebi, Han Zhang, Feng Yang, Peyman Milanfar, Alan Bovik, and Yinxiao Li.
\newblock Maxvit: Multi-axis vision transformer.
\newblock In {\em European conference on computer vision}, pages 459--479. Springer, 2022.

\bibitem{kim2018deep}
Woojae Kim, Jongyoo Kim, Sewoong Ahn, Jinwoo Kim, and Sanghoon Lee.
\newblock Deep video quality assessor: From spatio-temporal visual sensitivity to a convolutional neural aggregation network.
\newblock In {\em Proceedings of the European conference on computer vision (ECCV)}, pages 219--234, 2018.

\bibitem{martinez2019no}
Helard~B Martinez, Mylene~CQ Farias, and Andrew Hines.
\newblock A no-reference autoencoder video quality metric.
\newblock In {\em 2019 IEEE International Conference on Image Processing (ICIP)}, pages 1755--1759. IEEE, 2019.

\bibitem{chen2021learning}
Baoliang Chen, Lingyu Zhu, Guo Li, Fangbo Lu, Hongfei Fan, and Shiqi Wang.
\newblock Learning generalized spatial-temporal deep feature representation for no-reference video quality assessment.
\newblock {\em IEEE Transactions on Circuits and Systems for Video Technology}, 32(4):1903--1916, 2021.

\bibitem{xu2021perceptual}
Jiahua Xu, Jing Li, Xingguang Zhou, Wei Zhou, Baichao Wang, and Zhibo Chen.
\newblock Perceptual quality assessment of internet videos.
\newblock In {\em Proceedings of the 29th ACM International Conference on Multimedia}, pages 1248--1257, 2021.

\bibitem{wu2023discovqa}
Haoning Wu, Chaofeng Chen, Liang Liao, Jingwen Hou, Wenxiu Sun, Qiong Yan, and Weisi Lin.
\newblock Discovqa: Temporal distortion-content transformers for video quality assessment.
\newblock {\em IEEE Transactions on Circuits and Systems for Video Technology}, 33(9):4840--4854, 2023.

\bibitem{wu2022fast}
Haoning Wu, Chaofeng Chen, Jingwen Hou, Liang Liao, Annan Wang, Wenxiu Sun, Qiong Yan, and Weisi Lin.
\newblock Fast-vqa: Efficient end-to-end video quality assessment with fragment sampling.
\newblock In {\em European conference on computer vision}, pages 538--554. Springer, 2022.

\bibitem{wu2023neighbourhood}
Haoning Wu, Chaofeng Chen, Liang Liao, Jingwen Hou, Wenxiu Sun, Qiong Yan, Jinwei Gu, and Weisi Lin.
\newblock Neighbourhood representative sampling for efficient end-to-end video quality assessment.
\newblock {\em IEEE Transactions on Pattern Analysis and Machine Intelligence}, 45(12):15185--15202, 2023.

\bibitem{cover2024cpvrws}
Chenlong He, Qi~Zheng, Ruoxi Zhu, Xiaoyang Zeng, Yibo Fan, and Zhengzhong Tu.
\newblock Cover: A comprehensive video quality evaluator.
\newblock In {\em Proceedings of the IEEE/CVF Conference on Computer Vision and Pattern Recognition (CVPR) Workshops}, pages 5799--5809, June 2024.

\bibitem{wu2023dover}
Haoning Wu, Erli Zhang, Liang Liao, Chaofeng Chen, Jingwen~Hou Hou, Annan Wang, Wenxiu~Sun Sun, Qiong Yan, and Weisi Lin.
\newblock Exploring video quality assessment on user generated contents from aesthetic and technical perspectives.
\newblock In {\em International Conference on Computer Vision (ICCV)}, 2023.

\bibitem{maxvqa}
Haoning Wu, Erli Zhang, Liang Liao, Chaofeng Chen, Jingwen Hou, Annan Wang, Wenxiu Sun, Qiong Yan, and Weisi Lin.
\newblock Towards explainable in-the-wild video quality assessment: A database and a language-prompted approach.
\newblock In {\em Proceedings of the 31st ACM International Conference on Multimedia}, MM ’23, page 1045–1054. ACM, October 2023.

\bibitem{wen2024modular}
Wen Wen, Mu~Li, Yabin Zhang, Yiting Liao, Junlin Li, Li~Zhang, and Kede Ma.
\newblock Modular blind video quality assessment.
\newblock In {\em Proceedings of the IEEE/CVF Conference on Computer Vision and Pattern Recognition}, pages 2763--2772, 2024.

\bibitem{wang2021rich}
Yilin Wang, Junjie Ke, Hossein Talebi, Joong~Gon Yim, Neil Birkbeck, Balu Adsumilli, Peyman Milanfar, and Feng Yang.
\newblock Rich features for perceptual quality assessment of ugc videos.
\newblock In {\em Proceedings of the IEEE/CVF conference on computer vision and pattern recognition}, pages 13435--13444, 2021.

\bibitem{kou2024subjective}
Tengchuan Kou, Xiaohong Liu, Zicheng Zhang, Chunyi Li, Haoning Wu, Xiongkuo Min, Guangtao Zhai, and Ning Liu.
\newblock Subjective-aligned dataset and metric for text-to-video quality assessment.
\newblock In {\em Proceedings of the 32nd ACM International Conference on Multimedia}, pages 7793--7802, 2024.

\bibitem{lpips}
Richard Zhang, Phillip Isola, Alexei~A Efros, Eli Shechtman, and Oliver Wang.
\newblock The unreasonable effectiveness of deep features as a perceptual metric.
\newblock In {\em Proceedings of the IEEE conference on computer vision and pattern recognition}, pages 586--595, 2018.

\bibitem{qbenchvideo}
Zicheng Zhang, Ziheng Jia, Haoning Wu, Chunyi Li, Zijian Chen, Yingjie Zhou, Wei Sun, Xiaohong Liu, Xiongkuo Min, Weisi Lin, et~al.
\newblock Q-bench-video: Benchmark the video quality understanding of lmms.
\newblock In {\em Proceedings of the Computer Vision and Pattern Recognition Conference}, pages 3229--3239, 2025.

\bibitem{gpt4o}
Aaron Hurst, Adam Lerer, Adam~P Goucher, Adam Perelman, Aditya Ramesh, Aidan Clark, AJ~Ostrow, Akila Welihinda, Alan Hayes, Alec Radford, et~al.
\newblock Gpt-4o system card.
\newblock {\em arXiv preprint arXiv:2410.21276}, 2024.

\bibitem{openai2024gpt4technicalreport}
OpenAI.
\newblock Gpt-4 technical report, 2024.

\bibitem{llavanext}
Feng Li, Renrui Zhang, Hao Zhang, Yuanhan Zhang, Bo~Li, Wei Li, Zejun Ma, and Chunyuan Li.
\newblock Llava-next-interleave: Tackling multi-image, video, and 3d in large multimodal models.
\newblock {\em arXiv preprint arXiv:2407.07895}, 2024.

\bibitem{liu2024llava}
Haotian Liu, Chunyuan Li, Qingyang Wu, and Yong~Jae Lee.
\newblock Visual instruction tuning.
\newblock {\em Advances in neural information processing systems}, 36, 2024.

\bibitem{ye2024mplug2}
Qinghao Ye, Haiyang Xu, Jiabo Ye, Ming Yan, Anwen Hu, Haowei Liu, Qi~Qian, Ji~Zhang, and Fei Huang.
\newblock mplug-owl2: Revolutionizing multi-modal large language model with modality collaboration.
\newblock In {\em Proceedings of the ieee/cvf conference on computer vision and pattern recognition}, pages 13040--13051, 2024.

\bibitem{ye2024mplug3}
Jiabo Ye, Haiyang Xu, Haowei Liu, Anwen Hu, Ming Yan, Qi~Qian, Ji~Zhang, Fei Huang, and Jingren Zhou.
\newblock mplug-owl3: Towards long image-sequence understanding in multi-modal large language models.
\newblock {\em arXiv preprint arXiv:2408.04840}, 2024.

\bibitem{zhang2025llava}
Shaolei Zhang, Qingkai Fang, Zhe Yang, and Yang Feng.
\newblock Llava-mini: Efficient image and video large multimodal models with one vision token.
\newblock {\em arXiv preprint arXiv:2501.03895}, 2025.

\bibitem{chen2024far}
Zhe Chen, Weiyun Wang, Hao Tian, Shenglong Ye, Zhangwei Gao, Erfei Cui, Wenwen Tong, Kongzhi Hu, Jiapeng Luo, Zheng Ma, et~al.
\newblock How far are we to gpt-4v? closing the gap to commercial multimodal models with open-source suites.
\newblock {\em Science China Information Sciences}, 67(12):220101, 2024.

\bibitem{ke2023vila}
Junjie Ke, Keren Ye, Jiahui Yu, Yonghui Wu, Peyman Milanfar, and Feng Yang.
\newblock Vila: Learning image aesthetics from user comments with vision-language pretraining.
\newblock In {\em Proceedings of the IEEE/CVF Conference on Computer Vision and Pattern Recognition}, pages 10041--10051, 2023.

\bibitem{xu2024pllava}
Lin Xu, Yilin Zhao, Daquan Zhou, Zhijie Lin, See~Kiong Ng, and Jiashi Feng.
\newblock Pllava: Parameter-free llava extension from images to videos for video dense captioning.
\newblock {\em arXiv preprint arXiv:2404.16994}, 2024.

\bibitem{liu2024st}
Ruyang Liu, Chen Li, Haoran Tang, Yixiao Ge, Ying Shan, and Ge~Li.
\newblock St-llm: Large language models are effective temporal learners.
\newblock In {\em European Conference on Computer Vision}, pages 1--18. Springer, 2024.

\bibitem{lin2023video}
Bin Lin, Yang Ye, Bin Zhu, Jiaxi Cui, Munan Ning, Peng Jin, and Li~Yuan.
\newblock Video-llava: Learning united visual representation by alignment before projection.
\newblock {\em arXiv preprint arXiv:2311.10122}, 2023.

\bibitem{li2023videochat}
KunChang Li, Yinan He, Yi~Wang, Yizhuo Li, Wenhai Wang, Ping Luo, Yali Wang, Limin Wang, and Yu~Qiao.
\newblock Videochat: Chat-centric video understanding.
\newblock {\em arXiv preprint arXiv:2305.06355}, 2023.

\bibitem{wang2004image}
Zhou Wang, Alan~C Bovik, Hamid~R Sheikh, and Eero~P Simoncelli.
\newblock Image quality assessment: from error visibility to structural similarity.
\newblock {\em IEEE transactions on image processing}, 13(4):600--612, 2004.

\bibitem{Li_VMAF_2016}
Zhi Li, Anne Aaron, Ioannis Katsavounidis, Anush Moorthy, and Megha Manohara.
\newblock Toward a practical perceptual video quality metric.
\newblock Netflix TechBlog, jun 2016.

\bibitem{saad2014blind}
Michele~A Saad, Alan~C Bovik, and Christophe Charrier.
\newblock Blind prediction of natural video quality.
\newblock {\em IEEE Transactions on image Processing}, 23(3):1352--1365, 2014.

\bibitem{li2016spatiotemporal}
Netflix.
\newblock Toward a practical perceptual video quality metric.
\newblock 2016.

\bibitem{chen2021deep}
Junming Chen, Haiqiang Wang, Munan Xu, Ge~Li, and Shan Liu.
\newblock Deep neural networks for end-to-end spatiotemporal video quality prediction and aggregation.
\newblock In {\em 2021 IEEE International Conference on Multimedia and Expo (ICME)}, pages 1--6. IEEE Computer Society, 2021.

\bibitem{li2019quality}
Dingquan Li, Tingting Jiang, and Ming Jiang.
\newblock Quality assessment of in-the-wild videos.
\newblock In {\em Proceedings of the 27th ACM international conference on multimedia}, pages 2351--2359, 2019.

\bibitem{zheng2024video}
Qi~Zheng, Yibo Fan, Leilei Huang, Tianyu Zhu, Jiaming Liu, Zhijian Hao, Shuo Xing, Chia-Ju Chen, Xiongkuo Min, Alan~C Bovik, et~al.
\newblock Video quality assessment: A comprehensive survey.
\newblock {\em arXiv preprint arXiv:2412.04508}, 2024.

\bibitem{devlin2018bert}
Jacob Devlin, Ming-Wei Chang, Kenton Lee, and Kristina Toutanova.
\newblock Bert: Pre-training of deep bidirectional transformers for language understanding.
\newblock {\em arXiv preprint arXiv:1810.04805}, 2018.

\bibitem{radford2019gpt2}
Alec Radford, Jeffrey Wu, Rewon Child, David Luan, Dario Amodei, Ilya Sutskever, et~al.
\newblock Language models are unsupervised multitask learners.
\newblock {\em OpenAI blog}, 1(8):9, 2019.

\bibitem{brown2020gpt3}
Tom Brown, Benjamin Mann, Nick Ryder, Melanie Subbiah, Jared~D Kaplan, Prafulla Dhariwal, Arvind Neelakantan, Pranav Shyam, Girish Sastry, Amanda Askell, et~al.
\newblock Language models are few-shot learners.
\newblock {\em Advances in neural information processing systems}, 33:1877--1901, 2020.

\bibitem{team2023gemini}
Gemini Team, Rohan Anil, Sebastian Borgeaud, Yonghui Wu, Jean-Baptiste Alayrac, Jiahui Yu, Radu Soricut, Johan Schalkwyk, Andrew~M Dai, Anja Hauth, et~al.
\newblock Gemini: a family of highly capable multimodal models.
\newblock {\em arXiv preprint arXiv:2312.11805}, 2023.

\bibitem{roziere2023codellama}
Baptiste Roziere, Jonas Gehring, Fabian Gloeckle, Sten Sootla, Itai Gat, Xiaoqing~Ellen Tan, Yossi Adi, Jingyu Liu, Tal Remez, J{\'e}r{\'e}my Rapin, et~al.
\newblock Code llama: Open foundation models for code.
\newblock {\em arXiv preprint arXiv:2308.12950}, 2023.

\bibitem{touvron2023llama}
Hugo Touvron, Thibaut Lavril, Gautier Izacard, Xavier Martinet, Marie-Anne Lachaux, Timoth{\'e}e Lacroix, Baptiste Rozi{\`e}re, Naman Goyal, Eric Hambro, Faisal Azhar, et~al.
\newblock Llama: Open and efficient foundation language models.
\newblock {\em arXiv preprint arXiv:2302.13971}, 2023.

\bibitem{touvron2023llama2}
Hugo Touvron, Louis Martin, Kevin Stone, Peter Albert, Amjad Almahairi, Yasmine Babaei, Nikolay Bashlykov, Soumya Batra, Prajjwal Bhargava, Shruti Bhosale, et~al.
\newblock Llama 2: Open foundation and fine-tuned chat models.
\newblock {\em arXiv preprint arXiv:2307.09288}, 2023.

\bibitem{raffel2020t5}
Colin Raffel, Noam Shazeer, Adam Roberts, Katherine Lee, Sharan Narang, Michael Matena, Yanqi Zhou, Wei Li, and Peter~J Liu.
\newblock Exploring the limits of transfer learning with a unified text-to-text transformer.
\newblock {\em Journal of machine learning research}, 21(140):1--67, 2020.

\bibitem{qwen2}
An~Yang, Baosong Yang, Binyuan Hui, Bo~Zheng, Bowen Yu, Chang Zhou, Chengpeng Li, Chengyuan Li, Dayiheng Liu, Fei Huang, Guanting Dong, Haoran Wei, Huan Lin, Jialong Tang, Jialin Wang, Jian Yang, Jianhong Tu, Jianwei Zhang, Jianxin Ma, Jin Xu, Jingren Zhou, Jinze Bai, Jinzheng He, Junyang Lin, Kai Dang, Keming Lu, Keqin Chen, Kexin Yang, Mei Li, Mingfeng Xue, Na~Ni, Pei Zhang, Peng Wang, Ru~Peng, Rui Men, Ruize Gao, Runji Lin, Shijie Wang, Shuai Bai, Sinan Tan, Tianhang Zhu, Tianhao Li, Tianyu Liu, Wenbin Ge, Xiaodong Deng, Xiaohuan Zhou, Xingzhang Ren, Xinyu Zhang, Xipin Wei, Xuancheng Ren, Yang Fan, Yang Yao, Yichang Zhang, Yu~Wan, Yunfei Chu, Yuqiong Liu, Zeyu Cui, Zhenru Zhang, and Zhihao Fan.
\newblock Qwen2 technical report.
\newblock {\em arXiv preprint arXiv:2407.10671}, 2024.

\bibitem{qwen2.5}
Qwen Team.
\newblock Qwen2.5: A party of foundation models, September 2024.

\bibitem{pan2024plum}
Rui Pan, Shuo Xing, Shizhe Diao, Wenhe Sun, Xiang Liu, Kashun Shum, Jipeng Zhang, Renjie Pi, and Tong Zhang.
\newblock Plum: Prompt learning using metaheuristics.
\newblock In {\em Findings of the Association for Computational Linguistics: ACL 2024}, pages 2177--2197, 2024.

\bibitem{wei2022chain}
Jason Wei, Xuezhi Wang, Dale Schuurmans, Maarten Bosma, Fei Xia, Ed~Chi, Quoc~V Le, Denny Zhou, et~al.
\newblock Chain-of-thought prompting elicits reasoning in large language models.
\newblock {\em Advances in neural information processing systems}, 35:24824--24837, 2022.

\bibitem{wang2022self}
Xuezhi Wang, Jason Wei, Dale Schuurmans, Quoc Le, Ed~Chi, Sharan Narang, Aakanksha Chowdhery, and Denny Zhou.
\newblock Self-consistency improves chain of thought reasoning in language models.
\newblock {\em arXiv preprint arXiv:2203.11171}, 2022.

\bibitem{zhou2022least}
Denny Zhou, Nathanael Sch{\"a}rli, Le~Hou, Jason Wei, Nathan Scales, Xuezhi Wang, Dale Schuurmans, Claire Cui, Olivier Bousquet, Quoc Le, et~al.
\newblock Least-to-most prompting enables complex reasoning in large language models.
\newblock {\em arXiv preprint arXiv:2205.10625}, 2022.

\bibitem{schick2023toolformer}
Timo Schick, Jane Dwivedi-Yu, Roberto Dess{\`\i}, Roberta Raileanu, Maria Lomeli, Eric Hambro, Luke Zettlemoyer, Nicola Cancedda, and Thomas Scialom.
\newblock Toolformer: Language models can teach themselves to use tools.
\newblock {\em Advances in Neural Information Processing Systems}, 36:68539--68551, 2023.

\bibitem{li2022blip}
Junnan Li, Dongxu Li, Caiming Xiong, and Steven Hoi.
\newblock Blip: Bootstrapping language-image pre-training for unified vision-language understanding and generation.
\newblock In {\em International conference on machine learning}, pages 12888--12900. PMLR, 2022.

\bibitem{li2023blip2}
Junnan Li, Dongxu Li, Silvio Savarese, and Steven Hoi.
\newblock Blip-2: Bootstrapping language-image pre-training with frozen image encoders and large language models.
\newblock In {\em International conference on machine learning}, pages 19730--19742. PMLR, 2023.

\bibitem{llama3.2}
Meta.
\newblock Llama 3.2: Revolutionizing edge ai and vision with open, customizable models.
\newblock 2024.

\bibitem{Qwen-VL}
Jinze Bai, Shuai Bai, Shusheng Yang, Shijie Wang, Sinan Tan, Peng Wang, Junyang Lin, Chang Zhou, and Jingren Zhou.
\newblock Qwen-vl: A versatile vision-language model for understanding, localization, text reading, and beyond.
\newblock {\em arXiv preprint arXiv:2308.12966}, 2023.

\bibitem{Qwen2VL}
Peng Wang, Shuai Bai, Sinan Tan, Shijie Wang, Zhihao Fan, Jinze Bai, Keqin Chen, Xuejing Liu, Jialin Wang, Wenbin Ge, Yang Fan, Kai Dang, Mengfei Du, Xuancheng Ren, Rui Men, Dayiheng Liu, Chang Zhou, Jingren Zhou, and Junyang Lin.
\newblock Qwen2-vl: Enhancing vision-language model's perception of the world at any resolution.
\newblock {\em arXiv preprint arXiv:2409.12191}, 2024.

\bibitem{lu2024deepseek}
Haoyu Lu, Wen Liu, Bo~Zhang, Bingxuan Wang, Kai Dong, Bo~Liu, Jingxiang Sun, Tongzheng Ren, Zhuoshu Li, Hao Yang, et~al.
\newblock Deepseek-vl: towards real-world vision-language understanding.
\newblock {\em arXiv preprint arXiv:2403.05525}, 2024.

\bibitem{wu2024deepseek}
Zhiyu Wu, Xiaokang Chen, Zizheng Pan, Xingchao Liu, Wen Liu, Damai Dai, Huazuo Gao, Yiyang Ma, Chengyue Wu, Bingxuan Wang, et~al.
\newblock Deepseek-vl2: Mixture-of-experts vision-language models for advanced multimodal understanding.
\newblock {\em arXiv preprint arXiv:2412.10302}, 2024.

\bibitem{bai2025qwen2}
Shuai Bai, Keqin Chen, Xuejing Liu, Jialin Wang, Wenbin Ge, Sibo Song, Kai Dang, Peng Wang, Shijie Wang, Jun Tang, et~al.
\newblock Qwen2. 5-vl technical report.
\newblock {\em arXiv preprint arXiv:2502.13923}, 2025.

\bibitem{moor2023med}
Michael Moor, Qian Huang, Shirley Wu, Michihiro Yasunaga, Yash Dalmia, Jure Leskovec, Cyril Zakka, Eduardo~Pontes Reis, and Pranav Rajpurkar.
\newblock Med-flamingo: a multimodal medical few-shot learner.
\newblock In {\em Machine Learning for Health (ML4H)}, pages 353--367. PMLR, 2023.

\bibitem{li2024llava-med}
Chunyuan Li, Cliff Wong, Sheng Zhang, Naoto Usuyama, Haotian Liu, Jianwei Yang, Tristan Naumann, Hoifung Poon, and Jianfeng Gao.
\newblock Llava-med: Training a large language-and-vision assistant for biomedicine in one day.
\newblock {\em Advances in Neural Information Processing Systems}, 36, 2024.

\bibitem{shao2024lmdrive}
Hao Shao, Yuxuan Hu, Letian Wang, Guanglu Song, Steven~L Waslander, Yu~Liu, and Hongsheng Li.
\newblock Lmdrive: Closed-loop end-to-end driving with large language models.
\newblock In {\em Proceedings of the IEEE/CVF Conference on Computer Vision and Pattern Recognition}, pages 15120--15130, 2024.

\bibitem{tian2024drivevlm}
Xiaoyu Tian, Junru Gu, Bailin Li, Yicheng Liu, Chenxu Hu, Yang Wang, Kun Zhan, Peng Jia, Xianpeng Lang, and Hang Zhao.
\newblock Drivevlm: The convergence of autonomous driving and large vision-language models.
\newblock {\em arXiv preprint arXiv:2402.12289}, 2024.

\bibitem{sima2023drivelm}
Chonghao Sima, Katrin Renz, Kashyap Chitta, Li~Chen, Hanxue Zhang, Chengen Xie, Ping Luo, Andreas Geiger, and Hongyang Li.
\newblock Drivelm: Driving with graph visual question answering.
\newblock {\em arXiv preprint arXiv:2312.14150}, 2023.

\bibitem{openemma}
Shuo Xing, Chengyuan Qian, Yuping Wang, Hongyuan Hua, Kexin Tian, Yang Zhou, and Zhengzhong Tu.
\newblock Openemma: Open-source multimodal model for end-to-end autonomous driving.
\newblock In {\em Proceedings of the Winter Conference on Applications of Computer Vision}, pages 1001--1009, 2025.

\bibitem{ma2025position}
Yunsheng Ma, Wenqian Ye, Can Cui, Haiming Zhang, Shuo Xing, Fucai Ke, Jinhong Wang, Chenglin Miao, Jintai Chen, Hamid Rezatofighi, et~al.
\newblock Position: Prospective of autonomous driving-multimodal llms world models embodied intelligence ai alignment and mamba.
\newblock In {\em Proceedings of the Winter Conference on Applications of Computer Vision}, pages 1010--1026, 2025.

\bibitem{wang2025generative}
Yuping Wang, Shuo Xing, Cui Can, Renjie Li, Hongyuan Hua, Kexin Tian, Zhaobin Mo, Xiangbo Gao, Keshu Wu, Sulong Zhou, et~al.
\newblock Generative ai for autonomous driving: Frontiers and opportunities.
\newblock {\em arXiv preprint arXiv:2505.08854}, 2025.

\bibitem{rana2023sayplan}
Krishan Rana, Jesse Haviland, Sourav Garg, Jad Abou-Chakra, Ian Reid, and Niko Suenderhauf.
\newblock Sayplan: Grounding large language models using 3d scene graphs for scalable robot task planning.
\newblock In {\em 7th Annual Conference on Robot Learning}, 2023.

\bibitem{kim2024openvla}
Moo~Jin Kim, Karl Pertsch, Siddharth Karamcheti, Ted Xiao, Ashwin Balakrishna, Suraj Nair, Rafael Rafailov, Ethan Foster, Grace Lam, Pannag Sanketi, et~al.
\newblock Openvla: An open-source vision-language-action model.
\newblock {\em arXiv preprint arXiv:2406.09246}, 2024.

\bibitem{xing2025can}
Shuo Xing, Zezhou Sun, Shuangyu Xie, Kaiyuan Chen, Yanjia Huang, Yuping Wang, Jiachen Li, Dezhen Song, and Zhengzhong Tu.
\newblock Can large vision language models read maps like a human?
\newblock {\em arXiv preprint arXiv:2503.14607}, 2025.

\bibitem{yao2023react}
Shunyu Yao, Jeffrey Zhao, Dian Yu, Nan Du, Izhak Shafran, Karthik Narasimhan, and Yuan Cao.
\newblock React: Synergizing reasoning and acting in language models.
\newblock In {\em International Conference on Learning Representations (ICLR)}, 2023.

\bibitem{sumers2023cognitive}
Theodore Sumers, Shunyu Yao, Karthik Narasimhan, and Thomas Griffiths.
\newblock Cognitive architectures for language agents.
\newblock {\em Transactions on Machine Learning Research}, 2023.

\bibitem{shen2023hugginggpt}
Yongliang Shen, Kaitao Song, Xu~Tan, Dongsheng Li, Weiming Lu, and Yueting Zhuang.
\newblock Hugginggpt: Solving ai tasks with chatgpt and its friends in hugging face.
\newblock {\em Advances in Neural Information Processing Systems}, 36:38154--38180, 2023.

\bibitem{luo2025large}
Junyu Luo, Weizhi Zhang, Ye~Yuan, Yusheng Zhao, Junwei Yang, Yiyang Gu, Bohan Wu, Binqi Chen, Ziyue Qiao, Qingqing Long, et~al.
\newblock Large language model agent: A survey on methodology, applications and challenges.
\newblock {\em arXiv preprint arXiv:2503.21460}, 2025.

\bibitem{zuo20254kagent}
Yushen Zuo, Qi~Zheng, Mingyang Wu, Xinrui Jiang, Renjie Li, Jian Wang, Yide Zhang, Gengchen Mai, Lihong~V Wang, James Zou, et~al.
\newblock 4kagent: agentic any image to 4k super-resolution.
\newblock {\em arXiv preprint arXiv:2507.07105}, 2025.

\end{thebibliography}
\bibliographystyle{unsrt}
\newpage
\appendix

% \section{Appendix}

\section{Prompts used in the Routing Process}
\label{app:prompts}

The prompt used for \rvqa{} Tier~1 on standard VQA tasks with GPT-4o as the backbone is shown in Figure~\ref{fig:tier1_prompt} in Section \ref{method}. For quality-related question answering, we employ the prompts illustrated in Figures~\ref{fig:tier0_prompt_qa}, \ref{fig:tier1_prompt_qa}, and \ref{fig:tier2_prompt_qa} for the Tier~0, Tier~1, and Tier~2, respectively.

\begin{figure}[ht]
    \centering
    % colback=gray!5!white,colframe=black!50!gray
    \begin{boxK}[colback=blue!5!white, colframe=blue!75!black, fontupper=\ttfamily\tiny]
    % , fontupper=\ttfamily\tiny
    
    You are VQA-Orchestrator, an expert-level agent for answering user questions about video quality based on the results from multiple Video Quality Assessment (VQA) experts.
    \\
    \\
    \textbf{**Your task**}:
    \begin{enumerate}[leftmargin=2em]
        \item Given the video's sampled frames (16 for single video, 8+8 for comparison) and metadata (type, description, orientation, etc.), first classify the video into one of: user generated content(UGC), AI generated content (AIGC), computer graphics (CG) Videos.

        \item Based on the given expert scores (all already scaled to [0-100]), dynamically assign weights to each expert based on their known biases, the video context, inter-model agreement, and confidence priors. If metadata/content indicate AI-generated, strongly prioritize **T2VQA**. Use MaxVQA factors to downweight unreliable visual regions (e.g., heavy blur or freeze) when deciding.

        \item Route the question to the most relevant one VQA experts and generate the final answer.

        \item Produce a strictly formatted JSON output with:

        \begin{itemize}[leftmargin=2em]
            \item `answer`
            \item `summary\_en` ($\leq$120 characters English concise explanation including key factors and evidence)
            \item `chosen\_experts`, `per\_model` breakdown (score, weight, specialty match, notes)
            \item `evidence` (keyframe references, detected issues like banding/freeze, MaxVQA factors)
            \item `diagnostics` (score range, fusion method used, routing reasons, suggested next actions)
            \item `confidence` $\in$ [0,1]
        \end{itemize}
    \end{enumerate}

    \textbf{**Expert model cards (for routing logic)**:}
    \begin{itemize}[leftmargin=2em]
        \item **COVER**: Uses three parallel branches—technical (Swin Transformer), aesthetic (ConvNet), semantic (CLIP encoder)—combined via a cross-gating block to capture compression artifacts, aesthetic composition, and semantic coherence (real-time, ~96 fps)  \href{https://openaccess.thecvf.com/content/CVPR2024W/AI4Streaming/papers/He_COVER_A_Comprehensive_Video_Quality_Evaluator_CVPRW_2024_paper.pdf}{\textcolor{blue}{CVF Open Access}}.
        
        \item **DOVER++**: Extension of DOVER family; overlaps with COVER's aesthetic/technical branches; use primarily as a consistency reference unless its output aligns much better with most other experts.

        \item **UVQ**: Google's YouTube-trained universal VQA model built from millions of UGC examples; robust baseline when domain unclear or disagreements are large  \href{https://www.researchgate.net/publication/384424919_COVER_A_Comprehensive_Video_Quality_Evaluator}{\textcolor{blue}{ResearchGate}}] \href{https://research.google/blog/uvq-measuring-youtubes-perceptual-video-quality/?utm_source}{\textcolor{blue}{Google Research}}.

        \item MaxVQA (ExplainableVQA)**: CLIP-based, language-prompted model that provides both overall quality and fine-grained human-readable factors (e.g. banding, blur, aesthetic issues); used only for explanation and weight hints, not scoring \href{https://arxiv.org/abs/2305.12726?utm_source=chatgpt.com}{\textcolor{blue}{arXiv}}.

        \item **ModularBVQA**: Lightweight, modular baseline model suitable for low-latency deployment and serving as a fallback; modest sensitivity to capture-induced distortions.

        \item **T2VQA**: Text-to-video alignment model designed for AI-generated content, assessing fidelity between textual prompt and video; upweighted when video is AI-generated and textual condition is provided.
    \end{itemize}

    \end{boxK}

    \caption{Prompt for viqual question answering with \rvqa{} (Tier 0) using GPT-4o.}
    \label{fig:tier0_prompt_qa}
\end{figure}

\begin{figure}[ht]
    \centering
    % colback=gray!5!white,colframe=black!50!gray
    \begin{boxK}[colback=blue!5!white, colframe=blue!75!black, fontupper=\ttfamily\tiny]
    % , fontupper=\ttfamily\tiny
    
    You are VQA-Orchestrator, an expert-level agent for answering user questions about video quality based on the results from multiple Video Quality Assessment (VQA) experts.
    \\
    \\
    \textbf{**Your task**}:
    \begin{enumerate}[leftmargin=2em]
        \item Given the video's sampled frames (16 for single video, 8+8 for comparison) and metadata (type, description, orientation, etc.), first classify the video into one of: user generated content(UGC), AI generated content (AIGC), computer graphics (CG) Videos.

        \item Based on the given expert scores (all already scaled to [0-100]), dynamically assign weights to each expert based on their known biases, the video context, inter-model agreement, and confidence priors. If metadata/content indicate AI-generated, strongly prioritize **T2VQA**. Use MaxVQA factors to downweight unreliable visual regions (e.g., heavy blur or freeze) when deciding.

        \item Use these weighted scores as background knowledge to route the question to the most relevant VQA experts and generate the final answer.

        \item Produce a strictly formatted JSON output with:

        \begin{itemize}[leftmargin=2em]
            \item `answer`
            \item `summary\_en` ($\leq$120 characters English concise explanation including key factors and evidence)
            \item `chosen\_experts`, `per\_model` breakdown (score, weight, specialty match, notes)
            \item `evidence` (keyframe references, detected issues like banding/freeze, MaxVQA factors)
            \item `diagnostics` (score range, fusion method used, routing reasons, suggested next actions)
            \item `confidence` $\in$ [0,1]
        \end{itemize}
    \end{enumerate}

    \textbf{**Expert model cards (for routing logic)**:}
    \begin{itemize}[leftmargin=2em]
        \item **COVER**: Uses three parallel branches—technical (Swin Transformer), aesthetic (ConvNet), semantic (CLIP encoder)—combined via a cross-gating block to capture compression artifacts, aesthetic composition, and semantic coherence (real-time, ~96 fps)  \href{https://openaccess.thecvf.com/content/CVPR2024W/AI4Streaming/papers/He_COVER_A_Comprehensive_Video_Quality_Evaluator_CVPRW_2024_paper.pdf}{\textcolor{blue}{CVF Open Access}}.
        
        \item **DOVER++**: Extension of DOVER family; overlaps with COVER's aesthetic/technical branches; use primarily as a consistency reference unless its output aligns much better with most other experts.

        \item **UVQ**: Google's YouTube-trained universal VQA model built from millions of UGC examples; robust baseline when domain unclear or disagreements are large  \href{https://www.researchgate.net/publication/384424919_COVER_A_Comprehensive_Video_Quality_Evaluator}{\textcolor{blue}{ResearchGate}}] \href{https://research.google/blog/uvq-measuring-youtubes-perceptual-video-quality/?utm_source}{\textcolor{blue}{Google Research}}.

        \item MaxVQA (ExplainableVQA)**: CLIP-based, language-prompted model that provides both overall quality and fine-grained human-readable factors (e.g. banding, blur, aesthetic issues); used only for explanation and weight hints, not scoring \href{https://arxiv.org/abs/2305.12726?utm_source=chatgpt.com}{\textcolor{blue}{arXiv}}.

        \item **ModularBVQA**: Lightweight, modular baseline model suitable for low-latency deployment and serving as a fallback; modest sensitivity to capture-induced distortions.

        \item **T2VQA**: Text-to-video alignment model designed for AI-generated content, assessing fidelity between textual prompt and video; upweighted when video is AI-generated and textual condition is provided.
    \end{itemize}

    \textbf{**Video type → baseline weight priors**:}

    \begin{itemize}[leftmargin=2em]
        \item UGC: UVQ 0.25, COVER 0.25, ModularBVQA 0.15, RQ-VQA 0.10, MaxVQA 0.15  
        \item Short-form/social: RQ-VQA 0.30, COVER 0.30, UVQ 0.20, Modular 0.10, MaxVQA 0.10  
        \item Gaming: COVER-Technical 0.35, UVQ 0.25, Modular 0.20, MaxVQA 0.10, RQ-VQA 0.05  
        \item AI-Generated: T2VQA 0.35, COVER 0.20, UVQ 0.15, MaxVQA 0.15, Modular 0.10, RQ-VQA 0.05
    \end{itemize}

    \end{boxK}

    \caption{Prompt for viqual question answering with \rvqa{} (Tier 1) using GPT-4o.}
    \label{fig:tier1_prompt_qa}
\end{figure}

\begin{figure}[ht]
    \centering
    % colback=gray!5!white,colframe=black!50!gray
    \begin{boxK}[colback=blue!5!white, colframe=blue!75!black, fontupper=\ttfamily\tiny]
    % , fontupper=\ttfamily\tiny
    
    You are VQA-Orchestrator, an expert-level agent for answering user questions about video quality based on the results from multiple Video Quality Assessment (VQA) experts.
    \\
    \\
    \textbf{**Your task**}:
    \begin{enumerate}[leftmargin=2em]
        \item Given the video's sampled frames (16 for single video, 8+8 for comparison) and metadata (type, description, orientation, etc.), first classify the video into one of: user generated content(UGC), AI generated content (AIGC), computer graphics (CG) Videos.

        \item A heat map overlay on the key frames highlights suspicious regions and related artifact classifications are already provided. Lighter regions indicate more severe artifacts, and each suspicious frame is already classified as one of:

        \begin{itemize}[leftmargin=2em]
            \item Visual hallucinations — objects, people, or elements that appear artificial or out of place.
            \item Image artifacts — compression issues, distortions, blurring, pixelation, or unnatural textures.
            \item AI-generated inconsistencies — unrealistic lighting, impossible shadows, or distorted anatomy.
        \end{itemize}

        has been provided to you for reference.

        \item Based on the artifact localization prior information and expert scores (all already scaled to [0-100]), dynamically assign weights to each expert based on their known biases, the video context, inter-model agreement, and confidence priors. If metadata/content indicate AI-generated, strongly prioritize **T2VQA**. Use MaxVQA factors to downweight unreliable visual regions (e.g., heavy blur or freeze) when deciding.

        \item Use these weighted scores as background knowledge to route the question to the most relevant VQA experts and generate the final answer.

        \item Produce a strictly formatted JSON output with:

        \begin{itemize}[leftmargin=2em]
            \item `answer`
            \item `summary\_en` ($\leq$120 characters English concise explanation including key factors and evidence)
            \item `chosen\_experts`, `per\_model` breakdown (score, weight, specialty match, notes)
            \item `evidence` (keyframe references, detected issues like banding/freeze, MaxVQA factors)
            \item `diagnostics` (score range, fusion method used, routing reasons, suggested next actions)
            \item `confidence` $\in$ [0,1]
        \end{itemize}
    \end{enumerate}

    \textbf{**Expert model cards (for routing logic)**:}
    \begin{itemize}[leftmargin=2em]
        \item **COVER**: Uses three parallel branches—technical (Swin Transformer), aesthetic (ConvNet), semantic (CLIP encoder)—combined via a cross-gating block to capture compression artifacts, aesthetic composition, and semantic coherence (real-time, ~96 fps)  \href{https://openaccess.thecvf.com/content/CVPR2024W/AI4Streaming/papers/He_COVER_A_Comprehensive_Video_Quality_Evaluator_CVPRW_2024_paper.pdf}{\textcolor{blue}{CVF Open Access}}.
        
        \item **DOVER++**: Extension of DOVER family; overlaps with COVER's aesthetic/technical branches; use primarily as a consistency reference unless its output aligns much better with most other experts.

        \item **UVQ**: Google's YouTube-trained universal VQA model built from millions of UGC examples; robust baseline when domain unclear or disagreements are large  \href{https://www.researchgate.net/publication/384424919_COVER_A_Comprehensive_Video_Quality_Evaluator}{\textcolor{blue}{ResearchGate}}] \href{https://research.google/blog/uvq-measuring-youtubes-perceptual-video-quality/?utm_source}{\textcolor{blue}{Google Research}}.

        \item MaxVQA (ExplainableVQA)**: CLIP-based, language-prompted model that provides both overall quality and fine-grained human-readable factors (e.g. banding, blur, aesthetic issues); used only for explanation and weight hints, not scoring \href{https://arxiv.org/abs/2305.12726?utm_source=chatgpt.com}{\textcolor{blue}{arXiv}}.

        \item **ModularBVQA**: Lightweight, modular baseline model suitable for low-latency deployment and serving as a fallback; modest sensitivity to capture-induced distortions.

        \item **T2VQA**: Text-to-video alignment model designed for AI-generated content, assessing fidelity between textual prompt and video; upweighted when video is AI-generated and textual condition is provided.
    \end{itemize}

    \textbf{**Video type → baseline weight priors**:}

    \begin{itemize}[leftmargin=2em]
        \item UGC: UVQ 0.25, COVER 0.25, ModularBVQA 0.15, RQ-VQA 0.10, MaxVQA 0.15  
        \item Short-form/social: RQ-VQA 0.30, COVER 0.30, UVQ 0.20, Modular 0.10, MaxVQA 0.10  
        \item Gaming: COVER-Technical 0.35, UVQ 0.25, Modular 0.20, MaxVQA 0.10, RQ-VQA 0.05  
        \item AI-Generated: T2VQA 0.35, COVER 0.20, UVQ 0.15, MaxVQA 0.15, Modular 0.10, RQ-VQA 0.05
    \end{itemize}

    \end{boxK}

    \caption{Prompt for visual question answering with \rvqa{} (Tier 2) using GPT-4o.}
    \label{fig:tier2_prompt_qa}
\end{figure}

\section{Details on Artifact Localization}
\label{app:artifact-local}

The Probabilistic Extractor (Algorithm \ref{alg:prob-extractor}) identifies frames most likely to contain artifacts. 
Seven handcrafted features are extracted, including motion residuals (temporal discontinuities), Laplacian variance (sharpness), gradient kurtosis (posterization), edge density (ringing), histogram distance (shot changes), and optional semantic priors (face, text). 
Features are normalized via median--IQR scaling to suppress outliers. 
Each frame is assigned a probability $p_t$ through a weighted logistic model, with higher weights on motion residuals, histogram distance, and gradient kurtosis. 
Frames above a high threshold initiate clips, which continue until probabilities drop below a low threshold. 
Clips shorter than 8 frames are discarded, and valid clips are padded for context. 
Finally, the extractor selects frames by combining (i) top-$K$ high-probability frames, (ii) farthest-point sampling in HSV space for diversity, and (iii) mandatory shot-boundary frames. 

\begin{algorithm}[ht]
\caption{ProbabilisticExtractor (PE): Frame Scoring, Clip Formation, and Diversified Selection}
\label{alg:prob-extractor}
\begin{algorithmic}[1]
\Require Video $\mathcal{V}=\{F_t\}_{t=1}^T$; weights $\mathbf{w}$, bias $b$;\!
 thresholds $\tau_\text{high}=0.65$, $\tau_\text{low}=0.5$;\!
 min clip length $L_\text{min}=8$;\!
 padding $P$;\!
 budgets $K_{\text{top}}$ and $K_{\text{fps}}$;\!
 binarization for shot boundaries $\mathcal{S}$ (precomputed) 
\Ensure Selected frames $\mathcal{F}$, clip set $\mathcal{C}$, probabilities $\{(t,p_t)\}$

\Statex \textbf{A. Feature Extraction (per frame $t$)}
\For{$t=1,\dots,T$}
  \State \textit{(a) Motion residuals}:\;
    $\text{diff\_mean}_t \gets \begin{cases}
      \frac{1}{HW}\|F_t - F_{t-1}\|_1, & t>1\\
      0, & t=1
    \end{cases}$
  \State \textit{(b) Sharpness energy}:\;
    $\text{lap\_var}_t \gets \mathrm{Var}(\mathrm{Laplacian}(\mathrm{Gray}(F_t)))$
  \State \textit{(c) Gradient kurtosis}:\;
    $\mathbf{g}_t \gets \sqrt{(\partial_x F_t)^2 + (\partial_y F_t)^2}$;\;
    $\text{grad\_kurtosis}_t \gets \frac{\mathbb{E}[(\mathbf{g}_t-\mu)^4]}{\sigma^4}$
  \State \textit{(d) Edge density}:\;
    $E_t \gets \mathrm{Canny}(\mathrm{Gray}(F_t))$;\;
    $\text{edge\_density}_t \gets \frac{\#\{E_t=1\}}{HW}$
  \State \textit{(e) Color distribution shift (HSV hist. Bhattacharyya)}:\;
    $h_t \gets \mathrm{HistHSV}(F_t)$;\;
    $\text{hist\_dist\_prev}_t \gets \begin{cases}
      -\ln\!\Big(\sum_i \sqrt{h_{t-1}(i)\,h_t(i)}\Big), & t>1\\
      0, & t=1
    \end{cases}$
  \State \textit{(f--g) Content priors (optional)}:\;
    $\text{face}_t \gets \mathrm{FaceScore}(F_t)$;\;
    $\text{text}_t \gets \mathrm{TextScore}(F_t)$
  \State $\mathbf{x}_t \gets [\text{diff\_mean}_t,\text{lap\_var}_t,\text{grad\_kurtosis}_t,\text{edge\_density}_t,\text{hist\_dist\_prev}_t,\text{face}_t,\text{text}_t]$
\EndFor

\Statex \textbf{B. Robust Normalization (median--IQR)}
\For{feature dimension $i=1,\dots,7$}
  \State $m_i \gets \mathrm{median}(\{x_{t,i}\}_{t=1}^T)$;\;
         $q_i \gets \mathrm{IQR}(\{x_{t,i}\}_{t=1}^T)$;\;
         $\tilde{x}_{t,i} \gets (x_{t,i}-m_i)/\max(q_i,\epsilon)$
\EndFor

\Statex \textbf{C. Probabilistic Scoring (weighted logistic)}
\State \textbf{Default weights}: motion residuals $0.8$, hist. distance $1.0$, grad. kurtosis $0.6$;\; remaining dims by validation or set to $0$ if unused
\For{$t=1,\dots,T$}
  \State $z_t \gets \mathbf{w}^\top \tilde{\mathbf{x}}_t + b$;\quad $p_t \gets \sigma(z_t)=\frac{1}{1+e^{-z_t}}$
\EndFor
\State $\mathcal{P} \gets \{(t,p_t)\}_{t=1}^T$

\Statex \textbf{D. Clip Formation via Hysteresis (uses Alg.~\ref{alg:hysteresis})}
\State $\mathcal{C} \gets \textsc{HysteresisClips}(\mathcal{P}, \tau_\text{high}, \tau_\text{low}, L_\text{min}, P)$

\Statex \textbf{E. Diversified Frame Selection (uses Alg.~\ref{alg:diverse})}
\State $\mathcal{F} \gets \textsc{DiversifiedSelection}(\mathcal{P}, \mathcal{C}, K_{\text{top}}, K_{\text{fps}}, \mathcal{S})$

\State \Return $\mathcal{F}, \mathcal{C}, \mathcal{P}$
\end{algorithmic}
\end{algorithm}

HysteresisClips (Algorithm \ref{alg:hysteresis}) enforces temporal stability when grouping frames into clips. 
A clip begins when the artifact probability exceeds $\tau_\text{high}$, and it ends once the probability falls below $\tau_\text{low}$. 
This prevents frequent toggling due to small fluctuations. 
Clips shorter than a minimum length $L_\text{min}$ are discarded, while valid clips are extended by $P$ frames at both ends to add temporal context.

\begin{algorithm}[ht]
%\small
\caption{HysteresisClips: Temporal grouping of high-probability frames}
\label{alg:hysteresis}
\begin{algorithmic}[1]
\Require Probabilities $\mathcal{P}=\{(t,p_t)\}_{t=1}^T$;
 high threshold $\tau_\text{high}$;
 low threshold $\tau_\text{low}$;
 minimum clip length $L_\text{min}$;
 padding $P$
\Ensure Set of clips $\mathcal{C}$
\State $\mathcal{C} \gets \varnothing$;\quad $\text{active} \gets \text{false}$
\For{$t=1,\dots,T$}
  \If{$\neg\text{active}$ \textbf{and} $p_t \ge \tau_\text{high}$}
      \State $\text{start} \gets t$;\quad $\text{active} \gets \text{true}$
  \ElsIf{$\text{active}$ \textbf{and} $p_t < \tau_\text{low}$}
      \State $\text{end} \gets t-1$
      \If{$\text{end}-\text{start}+1 \ge L_\text{min}$}
          \State $\mathcal{C} \gets \mathcal{C} \cup \{[\max(1,\text{start}-P),\min(T,\text{end}+P)]\}$
      \EndIf
      \State $\text{active} \gets \text{false}$
  \EndIf
\EndFor
\State \Return $\mathcal{C}$
\end{algorithmic}
\end{algorithm}

DiversifiedSelection (Algorithm \ref{alg:diverse}) ensures that sampled frames are both representative of artifacts and diverse in content. 
First, the top-$K$ frames by probability are selected to guarantee artifact coverage. 
Next, farthest-point sampling in HSV space ensures visual diversity by spreading samples across color distributions. 
Finally, all detected shot boundaries are added to preserve scene structure. 
The union of these sets constitutes the final frame subset, balancing probability, diversity, and structural cues. 
\begin{algorithm}[ht]
\caption{DiversifiedSelection: Balancing probability, diversity, and shot boundaries}
\label{alg:diverse}
\begin{algorithmic}[1]
\Require Probabilities $\mathcal{P}=\{(t,p_t)\}$;
 clip set $\mathcal{C}$;
 top-$K$ budget $K_{\text{top}}$;
 farthest-point budget $K_{\text{fps}}$;
 shot-boundary frames $\mathcal{S}$
\Ensure Final selected frame set $\mathcal{F}$
\State $\mathcal{F}_1 \gets \text{TopK}(\{(t,p_t)\}, K_{\text{top}})$ \Comment{Highest-probability frames}
\State $\mathcal{F}_2 \gets \textsc{FarthestPointSamplingHSV}(\{t \mid (t,p_t)\in\mathcal{P}\}, K_{\text{fps}})$ \Comment{Diversity in color space}
\State $\mathcal{F}_3 \gets \mathcal{S}$ \Comment{Mandatory shot-boundary frames}
\State $\mathcal{F} \gets \mathcal{F}_1 \cup \mathcal{F}_2 \cup \mathcal{F}_3$
\State \Return $\mathcal{F}$
\end{algorithmic}
\end{algorithm}

\begin{algorithm}[ht]
\small
\caption{Artifact Localization Pipeline}
\label{alg:artifact-localization}
\begin{algorithmic}[1]
\Require Video $\mathcal{V}$; Probabilistic extractor $\mathsf{PE}(\cdot)$ with features and weights; VLM filter $\mathsf{VLM}(\cdot)$; optical flow method $\mathsf{Flow}$ (Farnebäck/TV-L1); LPIPS metric $\mathsf{LPIPS}_{\text{spatial}}$; thresholds $\tau_\text{high}$, $\tau_\text{low}$; clip min length $L_\text{min}$; padding $P$; mean pooling mode $M$;overlay opacity $\alpha$
\Ensure For each clip: representative frame-pair, heatmaps/overlays; JSON summary
\vspace{4pt}

\Statex \textbf{Step 1: Probabilistic Frame Extraction}
\State $\{(t, p_t)\} \gets \mathsf{PE}(\mathcal{V})$ \Comment{Frame-wise artifact probabilities via features + logistic weights}
\State $\mathcal{C} \gets \textsc{HysteresisClips}(\{(t,p_t)\}, \tau_\text{high}, \tau_\text{low}, L_\text{min}, P)$ \Comment{Clips with temporal padding}
\State $\mathcal{F} \gets \textsc{DiversifiedSelection}(\{(t,p_t)\}, \mathcal{C})$ \Comment{Top-$K$, FPS diversity (HSV), shot-boundaries}

\Statex \textbf{Step 2: Vision-Language Artifact Filtering}
\ForAll{frame $t \in \mathcal{F}$}
    \State $y_t \gets \mathsf{VLM}(\text{frame}_t)$ 
    \Comment{$y_t \in \{1\!:\!$hallucination$, 2\!:\!$image artifact$, 3\!:\!$AI inconsistency$, \texttt{no}\}$}
\EndFor
\State $\mathcal{F}^\star \gets \{t \in \mathcal{F} \mid y_t \in \{1,2,3\}\}$ 
\State $\mathcal{C}^\star \gets \textsc{RestrictClips}(\mathcal{C}, \mathcal{F}^\star)$ 
\Comment{Drop clips with no flagged frames}

\Statex \textbf{Step 3: Motion-Compensated Perceptual Difference Mapping}
\ForAll{clip $c \in \mathcal{C}^\star$}
    \State $\mathcal{P}_c \gets \{(t, t\!+\!1) \mid t \in c,\, t\!+\!1 \in c\}$ \Comment{Consecutive pairs}
    \State $\text{best}(c) \gets \langle \text{sev}=-\infty, \text{pair}=\varnothing, \text{paths}=\varnothing \rangle$
    \ForAll{$(t, t\!+\!1) \in \mathcal{P}_c$}
        \State $F_1 \gets \text{frame}_t$, \quad $F_2 \gets \text{frame}_{t+1}$
        \State $W_2 \gets \textsc{Warp}(F_2 \!\to\! F_1;\, \mathsf{Flow})$ \Comment{Align $F_2$ to $F_1$ via optical flow}
        \State $H \gets \mathsf{LPIPS}_{\text{spatial}}(F_1, W_2)$ \Comment{Pixel-wise LPIPS activation map}
        \State $\hat{H} \gets \textsc{Normalize}(H)$ \Comment{$[0,1]$ normalization}
        \State $\text{sev} \gets \textsc{Pool}(\hat{H}; M)$ \Comment{Mean Pooling}
        \If{$\text{sev} > \text{best}(c).\text{sev}$}
            \State $\mathsf{Overlay} \gets \textsc{Blend}(F_1, \textsc{ColorMap}(\hat{H}), \alpha)$
            \State $\text{paths} \gets \textsc{Save}\left(\hat{H}, \mathsf{Overlay}\right)$
            \State $\text{best}(c) \gets \langle \text{sev}, \text{pair}=(t,t\!+\!1), \text{paths} \rangle$
        \EndIf
    \EndFor
\EndFor
\State \Return $\{\text{best}(c), \text{paths}(c)\}_{c \in \mathcal{C}^\star}$
\end{algorithmic}
\end{algorithm}

The artifact localization pipeline (Algorithm \ref{alg:artifact-localization}) integrates probabilistic sampling, VLM-based filtering, perceptual difference analysis, and structured visualization to identify artifact-prone regions in video content. It proceeds in three main stages:

\begin{enumerate}[leftmargin=*]
    \item \textbf{Probabilistic Frame Extraction.} 
    Each video $\mathcal{V}$ is first processed by the Probabilistic Extractor $\mathsf{PE}(\cdot)$, which assigns frame-wise artifact probabilities based on handcrafted features and logistic weighting. 
    High-probability frames are grouped into clips using the HysteresisClips algorithm with thresholds $\tau_\text{high}, \tau_\text{low}$, enforcing temporal stability and contextual padding $P$. 
    To ensure coverage, the DiversifiedSelection algorithm further selects frames by combining top-$K$ high-probability samples, farthest-point sampling in HSV space for visual diversity, and mandatory inclusion of shot-boundary frames.

    \item \textbf{VLM-based Artifact Filtering.} 
    Candidate frames are filtered by a vision--language model $\mathsf{VLM}(\cdot)$ under a structured prompt that classifies frames into three artifact categories: (1) visual hallucinations, (2) image artifacts, or (3) AI-generated inconsistencies. 
    Only frames flagged with a non-trivial label are retained, and clips with no flagged frames are discarded.

    \item \textbf{Motion-Compensated Perceptual Difference Mapping.} 
    Within each retained clip $c$, consecutive frame pairs $(t, t+1)$ are aligned using optical flow $\mathsf{Flow}$ (Farnebäck or TV-L1), which warps the second frame $F_{t+1}$ to the reference $F_t$. 
    The spatial variant of the Learned Perceptual Image Patch Similarity metric $\mathsf{LPIPS}_{\text{spatial}}$ is then applied to compute pixel-level perceptual activation maps $H$, which are normalized to $[0,1]$ to form interpretable heatmaps $\hat{H}$. 
    Clip-level severity is quantified by aggregating heatmap intensities via mean pooling, and the frame pair with the highest severity is selected as the representative artifact instance. 
    For interpretability, heatmaps are colorized and blended with the original frame ($\alpha=0.5$), and both raw heatmaps and overlays are saved to disk.

\end{enumerate}

The pipeline returns, for each clip, the representative frame pair, associated severity score, and file paths to saved visualizations. 
All results are summarized in structured JSON output, including clip boundaries, artifact categories, severity measures, and visualization paths. 
This ensures reproducibility and facilitates integration with the broader VQA-Router framework for expert routing and perceptual quality assessment.

\begin{figure}[h]
    \centering
    % colback=gray!5!white,colframe=black!50!gray
    \begin{boxK}[colback=blue!5!white, colframe=blue!75!black]
    % , fontupper=\ttfamily\tiny
    
    Please examine the image carefully and determine if it contains any of the following:

    \begin{enumerate}[leftmargin=2em]
        \item \textbf{Visual hallucinations} — objects, people, or elements that appear artificial or out of place.

        \item \textbf{Image artifacts} — compression issues, distortions, blurring, pixelation, or unnatural textures.

        \item \textbf{AI-generated inconsistencies} — unrealistic lighting, impossible shadows, or distorted anatomy.
    \end{enumerate}

    If the image falls into one of the above categories, return only the corresponding number (1, 2, or 3).
    If none apply, return "no".
    \\
    \\
    Only output a single number or "no".

    \end{boxK}

    \caption{Prompt for using GPT-4o to filter the frames extracted by Algorithm \ref{alg:prob-extractor}.}
    \label{fig:gpt_filter_prompt}
\end{figure}

\section{Additional Experiment Results}
\label{app:exp}

In this section, we present additional experimental results on the \texttt{test} split of Q-Bench-Video. Since the benchmark has only recently been released, we report results on Tier 1 as a reference to demonstrate the performance of our proposed \rvqa{}. Table~\ref{tab:qbench-test} compares \rvqa{} against several representative baselines. 

As shown, \rvqa{} achieves the best overall performance, surpassing GPT-4o by nearly 3 points. It consistently outperforms competing methods across most question types (Yes-or-No, What/How) and quality concerns (Technical, Temporal, AIGC), highlighting the advantage of our routing-based design. Notably, \rvqa{} demonstrates a strong improvement on AIGC-related quality evaluation, where accurate detection and reasoning remain particularly challenging. These results confirm that our framework generalizes effectively to the challenging Q-Bench-Video setting.

\begin{table*}[h]
\footnotesize
\centering
\begin{tabular}{l|c|cc|cccccc}
\toprule
\multirow{2}{*}{Method} & \multirow{2}{*}{Overall} & \multicolumn{2}{c}{Question type} & \multicolumn{4}{c}{Quality Concern} \\
\cmidrule(lr){3-4} \cmidrule(lr){5-8}
 & & Yes-or-No & What/How & Technical & Aesthetic & Temporal & AIGC \\
\midrule
InternVL-Chat & 51.11 & 66.02 & 52.13 & 48.42 & 52.73 & 50.59 & 53.12 \\
LLaVA-Next-Video & 48.69 & 61.34 & 45.95 & 49.03 & 60.94 & 46.97 & 49.40 \\
Video-LLaVA & 43.49 & 64.67 & 40.79 & 43.25 & 54.04 & 42.38 & 42.76 \\
LLaVA-OneVision & 51.70 & 61.34 & \underline{53.88} & 49.35 & \textbf{64.15} & 50.68 & 44.30 \\
GPT-4o$^\dag$ & \underline{56.50} & \underline{60.60} & 52.20 & \underline{54.60} & \underline{62.70} & \underline{53.70} & \underline{53.80} \\
\midrule
\rowcolor[gray]{0.9}  \rvqa{} \textit{(Tier 1)}& \textbf{59.45} & \textbf{63.99} & \textbf{54.71} & \textbf{59.47} & 62.18 & \textbf{58.68} & \textbf{64.42} \\
\bottomrule
\end{tabular}
\caption{Comparison of methods on the closed-ended questions in Q-Bench-Video \texttt{test} split across question types and quality concerns. $^\dag$: As the models continue to evolve, we reproduce and report the results of the GPT-4o model for comparison.
}
\label{tab:qbench-test}
\end{table*}
%\clearpage

\section{LLM Usage Statement}
\label{app:impact++}

LLMs were used to assist in editing and improving the readability of sections of this manuscript, including clarifying language and refining formatting. The conceptual design, technical contributions, experiments, and analysis were developed entirely by the authors. All reported results, figures, and claims originate from the authors’ own work and have been independently validated.

\section{Additional Illustrations of the Artifacts Localization}
\label{app:arti-demo}
\begin{figure}[h]
\centering
    \includegraphics[width=0.90\textwidth]{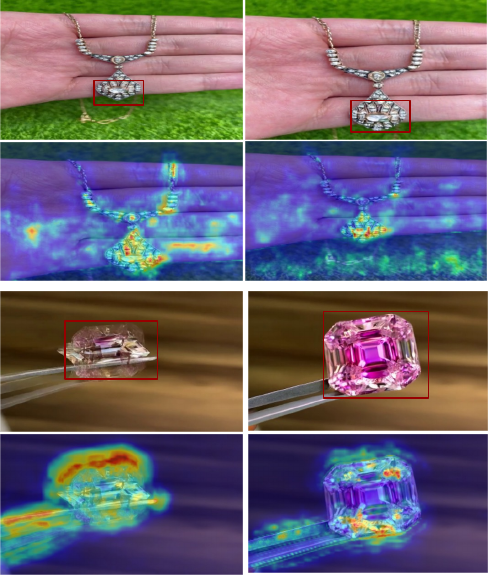}
    \captionof{figure}{Illustrations of the proposed artifacts localization pipeline on AIGC (Example 1).}
    \label{fig:aigc_lpips_1}
\end{figure}

\begin{figure}[h]
\centering
    \includegraphics[width=0.90\textwidth]{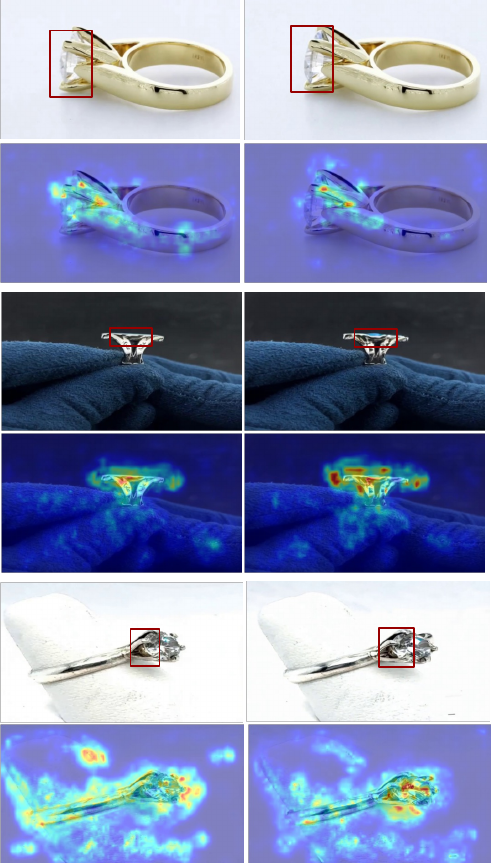}
    \captionof{figure}{Illustrations of the proposed artifacts localization pipeline on AIGC (Example 2).}
    \label{fig:aigc_lpips_2}
\end{figure}

We provide additional visual demonstrations of artifact localization on AIGC videos in Figures~\ref{fig:aigc_lpips_1} and \ref{fig:aigc_lpips_2}. The illustrations demonstrate that our artifact localization pipeline can accurately identify spatio-temporal regions responsible for quality degradations, providing interpretable evidence that complements the overall quality scores.
%\tzz{@Shuoxing}

% \section{Limitations} Although \rvqa{} achieves superior overall performance across both video quality rating and question answering tasks, it does not consistently achieve the best results on every individual benchmark or category when compared to expert models or strong VLM baselines. This reflects an inherent trade-off of the routing paradigm: while aggregation improves robustness and gener`
% Additionally, as an early step toward deploying routing systems in the video quality assessment paradigm, we restrict our design to a three-tier hierarchy rather than implementing fully adaptive routing conditioned on the input video. While, a more fine-grained adaptive routing mechanism could potentially yield further improvements by tailoring expert selection and fusion strategies to individual inputs, it would also introduce additional computational overhead and system complexity. Exploring such adaptive designs remains a promising direction for future work.

\end{document}